%% file: main.tex
\documentclass[10pt,twocolumn,letterpaper]{article}

 \usepackage[pagenumbers]{cvpr} 

\input{preamble}

\usepackage[accsupp]{axessibility}  
\usepackage{graphicx}
\usepackage{amsmath}
\usepackage{amssymb}
\usepackage{mathabx}
\usepackage{booktabs}
\usepackage{float}
\usepackage{algorithm,algorithmic}
\usepackage{xcolor}
\usepackage{multirow}
\usepackage{array}
\newcolumntype{M}[1]{>{\centering\arraybackslash}m{#1}}
\newcolumntype{L}[1]{>{\flushleft\arraybackslash}m{#1}}

\usepackage{caption} 

\definecolor{cvprblue}{rgb}{0.21,0.49,0.74}
\usepackage[pagebackref,breaklinks,colorlinks,citecolor=cvprblue]{hyperref}

\def\paperID{9606} 
\def\confName{CVPR}
\def\confYear{2024}

\title{Towards Memorization-Free Diffusion Models}

\author{
 Chen Chen \ \ \ \ \ \ \ \ \ Daochang Liu \ \ \ \ \ \ \ \ \ Chang Xu
 \and School of Computer Science, Faculty of Engineering, The University of Sydney
 \and {\tt\small \{cche0711@uni., daochang.liu@, c.xu@\}sydney.edu.au}
}
\begin{document}

\maketitle
\input{sec/0_abstract_camera_ready}    
\input{sec/1_intro_camera_ready}
\input{sec/2_related_work_camera_ready}

\input{sec/3_preliminaries_camera_ready}

\input{sec/4_memorization_camera_ready}
\input{sec/5_methodology_camera_ready}

\input{sec/6_evaluation_camera_ready}

\input{sec/7_conclusions_camera_ready}
{
    \small
    \bibliographystyle{ieeenat_fullname}
    \bibliography{main}
}
\input{sec/X_suppl}
\end{document}

%% file: preamble.tex
\usepackage[dvipsnames]{xcolor}


%% file: sec/0_abstract_camera_ready.tex
\begin{abstract}
Pretrained diffusion models and their outputs are widely accessible due to their exceptional capacity for synthesizing high-quality images and their open-source nature. The users, however, may face litigation risks owing to the models' tendency to memorize and regurgitate training data during inference. To address this, we introduce Anti-Memorization Guidance (AMG), a novel framework employing three targeted guidance strategies for the main causes of memorization: image and caption duplication, and highly specific user prompts. Consequently, AMG ensures memorization-free outputs while maintaining high image quality and text alignment, leveraging the synergy of its guidance methods, each indispensable in its own right. AMG also features an innovative automatic detection system for potential memorization during each step of inference process, allows selective application of guidance strategies, minimally interfering with the original sampling process to preserve output utility. We applied AMG to pretrained Denoising Diffusion Probabilistic Models (DDPM) and Stable Diffusion across various generation tasks. The results demonstrate that AMG is the first approach to successfully eradicates all instances of memorization with no or marginal impacts on image quality and text-alignment, as evidenced by FID and CLIP scores.
\vspace{-0.7cm}
\end{abstract}

%% file: sec/1_intro_camera_ready.tex
\section{Introduction}
\label{sec:intro}
Diffusion models \cite{Diffusion2015, ddpm, iddpm_2021_icml} have attracted substantial interest, given their superiority in terms of diversity, fidelity, scalability \cite{DALLE2} and controllability \cite{GLIDE} over previous generative models including VAEs \cite{VAE}, normalizing flows \cite{NF}, and GANs \cite{GAN, pggan, stylegan, stylegan2}.
With guidance techniques \cite{adm, cfg}, diffusion models can be further improved by the strategical diversity-fidelity trade-off.
State-of-the-art diffusion models trained on vast web-scale datasets are widespreadly used and have seen deployment at a commercial scale \cite{sd_2022_cvpr, midjourney, imagen}.

Such widespread adoption, however, has significantly heightened the litigation risks for companies using these models, particularly due to allegations that the models \textbf{memorize and reproduce training data during inference} without informing the data owners and the users of diffusion models. 
This potentially violates copyright laws and introduces ethical dilemmas, further complicated by the fact that the extensive size of training sets impedes detailed human review, leaving the intellectual property rights of the data sources largely undetermined.
An ongoing example is that a legal action contends that Stable Diffusion is \textit{a 21st-century collage tool that remixes the copyrighted works of millions of artists whose work was used as training data} \cite{lawsuit}.

\begin{figure}[t]
\hfill
\includegraphics[width=1.0\linewidth]{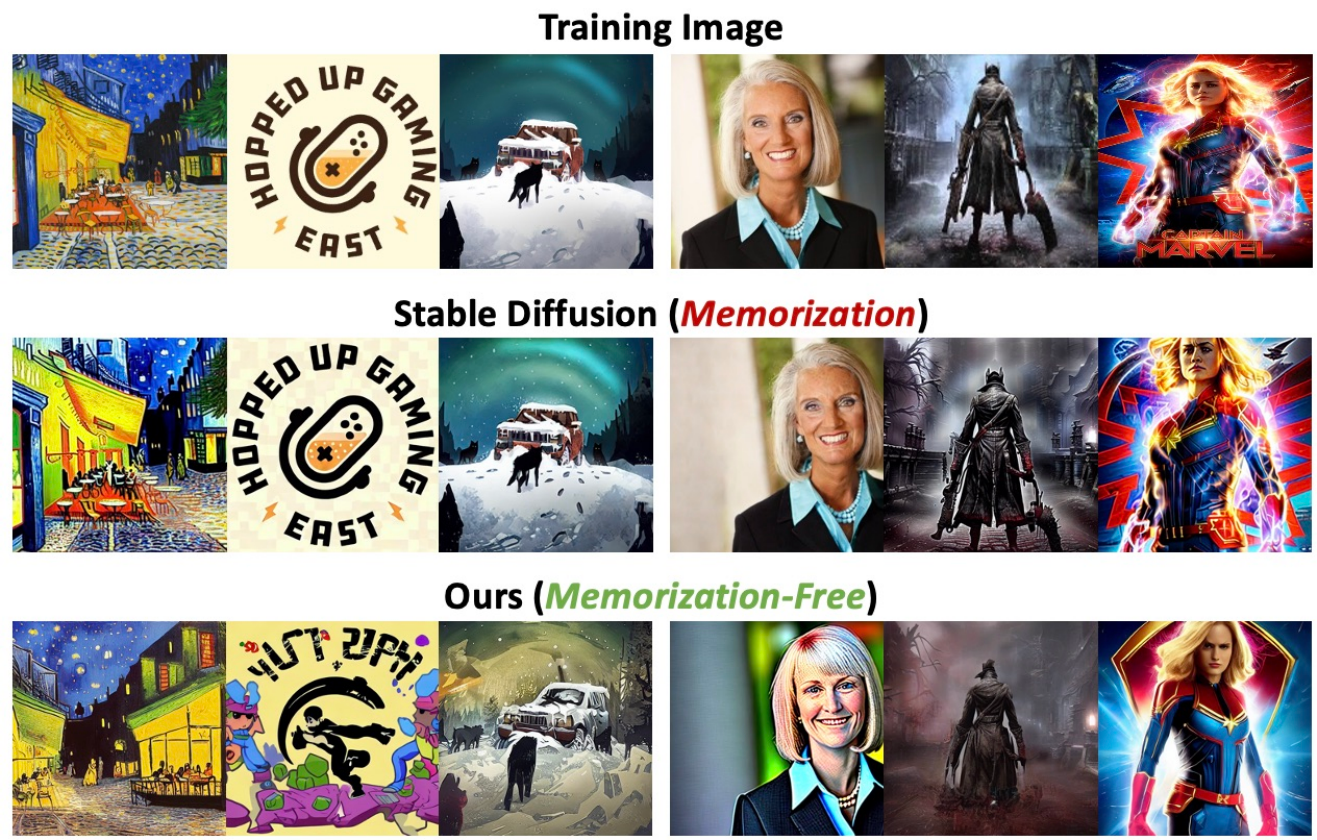}
\caption{Stable Diffusion's capacity to memorize training data, manifested as pixel-level memorization (left) and object-level memorization (right). Our approach successfully guides pretrained diffusion models to produce memorization-free outputs.}
\label{fig:sd-teasor}
\vspace{-0.7cm}
\end{figure}
Prior studies \cite{carlini_2023_usenix, somepalli_2023_cvpr, somepalli_2023_neurips} have observed memorization in pretrained diffusion models, particularly during unconditional CIFAR-10 \cite{cifar10} and text-conditional LAION dataset \cite{LAION5B} generations. While previous research proposed strategies to reduce memorization, these often lead to only modest improvements and fail to fully eliminate the issue. The effectiveness often come with reduced output quality and text-alignment \cite{somepalli_2023_neurips}, the need for retraining models \cite{carlini_2023_usenix}, and extensive manual intervention \cite{kumari_2023_iccv}. Moreover, these strategies lack an automated way to differentiate potential memorization cases for targeted mitigation. For example, \cite{kumari_2023_iccv} relies on a predefined list of text prompts prone to causing memorization, and \cite{somepalli_2023_neurips} applies randomization mechanisms uniformly without distinguishing between scenarios.

In this paper, we undertake the following systematic efforts to address the issue of memorization.
\textit{Firstly}, we have identified and detailed the primary causes of memorization, pinpointing image and text duplication in training datasets, along with the high specificity of user prompts for text conditioning, as key contributors.
\textit{Secondly}, we propose a novel unified framework, Anti-Memorization Guidance (AMG), which comprises three distinct guidance strategies, namely, \textit{desspecification guidance} ($G_{spe}$), \textit{caption deduplication guidance} ($G_{dup}$), and \textit{dissimilarity guidance} ($G_{sim}$), with each meticulously crafted to address one of these identified causes. 
Each strategy within AMG effectively guides generations away from memorized training images, offering unique benefits. $G_{spe}$ and $G_{dup}$ excel in maximally preserving the quality of generated images, while $G_{sim}$ provides a definitive assurance against memorization. The absence of any one of these strategies would compromise the delicate balance between privacy and utility, underscoring the indispensability of each of the three methods in the framework.
To further enhance the privacy-utility trade-off, AMG features an automatic detection mechanism that continuously assesses the similarity between the current prediction and its nearest training data during the inference process to identify potential instances of memorization. This allows AMG to apply guidance selectively rather than uniformly, ensuring that the original sampling process of the pretrained diffusion model is maximally preserved.

We conducted experiments with AMG on pretrained Denoising Diffusion Probabilistic Models (DDPM) and Stable Diffusion, spanning various generation tasks such as unconditional, class-conditional, and text-conditional generations. The outcomes, both qualitative and quantitative, demonstrate that AMG is the first method that effectively eradicates all memorization instances with minimal impact on image quality and text-alignment. 
In summary, our contributions through AMG are multifaceted and significant:
1) AMG introduces three guidance strategies, each meticulously designed to address one of the primary causes of memorization, providing a comprehensive solution that effectively balances privacy and utility.
2) AMG is equipped with an automatic detection system for potential memorization during each step of the inference process. This allows for the selective application of guidance strategies, maximizing the preservation of output utility.
3) Expanding upon previous research that focused only on unconditional and text-conditional generations, our study is the first to identify and address memorization in class-conditional diffusion model generations, and is the first that successfully achieves memorization-free generations with minimal compromise on image quality and text-alignment.

%% file: sec/2_related_work_camera_ready.tex
\vspace{-0.34cm}
\section{Related Work}
\label{sec:related_work}
\vspace{-0.3cm}
\textbf{Memorization in Diffusion Models}
has received increased scrutiny over the past year.
\cite{somepalli_2023_cvpr} found that pretrained Stable Diffusions and unconditional DDPMs trained on small datasets like CelebA \cite{celeba} and Oxford Flowers \cite{oxfordflowers} often replicate training data. \cite{carlini_2023_usenix} reported memorization in pretrained DDPMs on CIFAR-10.
Our work focus on pretrained diffusion models, which are extensively used and directly expose their users to litigation risks. We also pioneer in studying class-conditional model memorization, proposing a unified framework that successfully eradicates memorization in various generation tasks including unconditional, class-conditional, and text-conditional generations.

\textbf{Mitigation Strategies}.
\textit{Training data deduplication}, initially effective in language models \cite{language_dedup, language_dedup2}, was adapted for diffusion models \cite{carlini_2023_usenix}, who removed 5,275 similar images from CIFAR-10 and retrained the model, achieving a reduction in memorization. Yet, it offers limited improvement and requires retraining the entire model, which is computationally intensive, especially for advanced diffusion models with large datasets.
\textit{Concept ablation} \cite{kumari_2023_iccv}, implemented for Stable Diffusion, involved fine-tuning the pre-trained model on two sets of text-image pairs (one prone to memorization and the other not), curated using ChatGPT-generated paraphrases, to minimize their output disparity. 
While effective, this method demands extensive manual effort and relies heavily on the crafted prompts' quality. Also, it assumes the availability of a predefined list of memorization-prone prompts, which is unrealistic in many cases.
\textit{Randomizing text conditioning} \cite{somepalli_2023_neurips} during training or inference can also reduce memorization but has limitations. It mitigates, rather than fully prevents, memorization and lacks guaranteed effectiveness in untested scenarios. It results in significant declines in CLIP \cite{clip} score, indicating a poorly balanced trade-off between reducing memorization and maintaining text-alignment of generated images. Furthermore, its uniform application across all text conditions, without considering their potential for causing memorization, further diminishes the images' practical value.
Our approach, AMG, adopts a distinct strategy. Instead of altering text conditions to indirectly affect image generation, we aim to directly modify the generated image, thereby providing a guarantee of reduced similarity and addressing the issue of memorization more effectively.
Moreover, AMG uses real-time similarity metrics to selectively apply guidance to likely duplicates during inference, ensuring a targeted approach that leaves unaffected cases unaltered, in contrast to the indiscriminate application of \textit{randomization techniques}, also bypassing the need for manual crafting of \textit{concept ablation}.

\textbf{Other Privacy-Preservation Strategies} encompass the adoption of Differential Privacy (DP) \cite{dp} in training generative models \cite{dpgan, dp-gan-dpac, dpdm}, primarily through the use of the differentially-private stochastic gradient descent (DP-SGD) algorithm \cite{dpsgd}. 
Although effective on smaller datasets like MNIST and CelebA, \cite{carlini_2023_usenix} noted that implementing DP-SGD in diffusion models tends to result in divergence during training on datasets of CIFAR-10 scale. The feasibility of applying DP-SGD to even larger datasets, such as LAION, remain unexplored.
In addition, methods such as dataset distillation \cite{wang2018dataset, liu2022dataset} offer a means to prevent raw data being directly used in the training of generative models, thereby aiding in privacy preservation. Yet, there has been no exploration of these methods on the LAION dataset to date.

%% file: sec/3_preliminaries_camera_ready.tex
\vspace{-0.2cm}
\section{Preliminaries}
\label{sec:preliminaries}
\vspace{-0.2cm}
\subsection{Diffusion Models}
\label{sec:DM}
\textbf{Denoising Diffusion Probabilistic Models (DDPMs)} \cite{ddpm, iddpm_2021_icml} consist of two processes, firstly, a forward process is required to gradually add Gaussian noise to an image sampled from a real-data distribution $x_0 \sim q(x)$ over $T$ timesteps such that $x_T \sim \mathcal{N}(0, \mathbf{I})$.
The Diffusion Kernel then enables sampling $x_t$ at arbitrary timestep $t$ in a closed form:
\begin{equation}
q(x_t|x_0) = \mathcal{N}(x_t; \sqrt{\bar{\alpha}_t} x_0, (1 - \bar{\alpha}_t)\mathbf{I})
  \label{eq:pre3}
\end{equation}
where $x_t$ is the noised version of $x_0$ at timestep t, $\alpha_t = 1 - \beta_t$ is the noise schedule controls the amount of noise injected into the data at each step, and $\bar{\alpha}_t = \prod_{s=1}^{t} \alpha_s$.
Then, in the backward process, generative modelling can be realized by learning a denoiser network $\epsilon_{\theta}$ to predict the noise $\epsilon_t$ instead of the image $x_{t-1}$ at any arbitrary step $t$:
\begin{equation}
\mathcal{L} = \mathbb{E}_{t \in [1,T],\epsilon \sim \mathcal{N}(0, \mathbf{I})} [\left\| \epsilon_t - \epsilon_{\theta} (x_t, t) \right\|_2^2]
  \label{eq:pre6}
\end{equation}
where the denoiser network can be easily reformulated as a conditional generative model $\epsilon_{\theta} (x_t, y, t)$ by incorporating additional class or text conditioning $y$.

\textbf{Score-based formulation} \cite{score_sde} aims to construct a continuous time diffusion process, where $t \in [0,T]$ is continuous. The reverse processes can be formulated as:
\begin{equation}
dx_t = \left[ -\frac{1}{2} \beta(t) x_t - \beta(t) \nabla_{x_t} \log q_t(x_t) \right] dt + \sqrt{\beta(t)} d \bar{w}_t
  \label{eq:pre8}
\end{equation}
where $\beta(t)$ is a time-dependent function that allows different step sizes $\beta_t = \beta(t) \Delta t$ along the process $t$. A denoiser network $\nabla_{x_t} \log p_\theta(x_t)$ is then learned to approximate the score function $\nabla_{x_t} \log q_t(x_t)$ in the reverse process using a denoising score matching objective, which can be derived to be the same objective as in \cref{eq:pre6}, leveraging the connection between diffusion models and score matching:
\begin{equation}
\nabla_{x_t} \log p_{\theta}(x_t) = - \frac{1}{\sqrt{1 - \alpha_t}} \epsilon_{\theta}(x_t)
  \label{eq:pre9}
\end{equation}

\subsection{Guidance in Diffusion Models}
\label{sec:guidance}
Classifier guidance (CG) and classifier-free guidance (CFG) are methods used in diffusion models to steer image generation towards higher likelihood outcomes as determined by an explicit or implicit classifier $p_\phi(y|x_t)$.
In the score-based framework \cite{score_sde}, CG and CFG involve learning the gradient of the log probability for the conditional model, $\nabla_{x_t} \log p_{\theta}(x_t | y)$, rather than the score of unconditional model, $\nabla_{x_t} \log p_{\theta}(x_t)$. The conditional score can be easily derived using Bayes' rule as the sum of the unconditional score and the gradient of the log classifier probability:
\begin{equation}
\nabla_{x_t} \log p_{\theta}(x_t | y) = \nabla_{x_t} \log p_{\theta}(x_t) + \nabla_{x_t} \log p_{\phi}(y | x_t)
  \label{eq:pre10}
\end{equation}
\textbf{Classifier Guidance (CG)} \cite{adm} involves training an explicit classifier $p_{\phi}(y | x_t)$ on perturbed images $x_t$ and then employing its gradients $\nabla_{x_t} \log p_{\phi}(y | x_t)$ to direct the diffusion sampling process towards a class label $y$. Inserting \cref{eq:pre9} into \cref{eq:pre10}, \cite{adm} shows a new epsilon prediction $\hat{\epsilon}$ corresponds to the score presented in \cref{eq:pre10}:
\begin{equation}
\hat{\epsilon} := \epsilon_{\theta}(x_t) - \sqrt{1 - \bar{\alpha}_t} \nabla_{x_t} \log p_{\phi}(y|x_t)
  \label{eq:pre11}
\end{equation}
\textbf{Classifier-Free Guidance (CFG)} \cite{cfg} eliminates the need of an explicit classifier for computing \cref{eq:pre10}. 
It requires concurrent training on conditional and unconditional objectives. 
At inference, the epsilon prediction is linearly directed towards the conditional prediction and away from the unconditional, and $s_0 > 1$ controls the degree of adjustment:
\vspace{-0.3cm}
\begin{equation}
\hat{\epsilon} \leftarrow \epsilon_{\theta}(x_t) + s_0 \cdot (\epsilon_{\theta}(x_t, y)-\epsilon_{\theta}(x_t)).
  \label{eq:pre12}
\end{equation}

%% file: sec/4_memorization_camera_ready.tex
\section{Memorization in Diffusion Models}
\label{sec:memorization}
\label{sec:mem-sub1}
Memorization in generative models is identified when generated images exhibit extreme similarity to certain training samples.
The strictest definition of memorization relates to high pixel-level similarity, often qualitatively represented as the generated image being near-copies of training samples as in \cref{fig:cifar-250} and left of \cref{fig:sd-teasor}.
To quantify this similarity, the negative normalized Euclidean L2-norm distance (nL2) is employed as a pixel-level metric \cite{carlini_2023_usenix}. For a generated representation $\hat{x}_0$, this involves first identifying its nearest neighbor $n_0$ using the $\ell_2$ norm, and then normalizing the norm as follows:
\vspace{-0.3cm}
\begin{equation}
\sigma_t = - \frac{\ell_2(\hat{x}_0,n_0)}{\alpha \cdot \frac{1}{k} \sum_{z_0 \in S_{\hat{x}_0}} \ell_2(\hat{x}_0,z_0)}
  \label{eq:3}
\end{equation}
where $S_{\hat{x}_0}$ is a set of $k=50$ nearest neighbors of $\hat{x}_0$, and $\alpha$ is a scaling constant with a default value of $0.5$.

A broader definition of memorization encompasses reconstructive memory in diffusion models, where the models reassemble various elements from memorized training images, such as foreground and background objects. These reconstructions might include transformations like shifting, scaling, or cropping. Consequently, the reconstructed outputs do not necessarily match any training image on a pixel-by-pixel basis, yet they exhibit a high degree of similarity to certain training images at the object level. 
In right section of \cref{fig:sd-teasor}, we observe that the outputs generated by Stable Diffusion are not pixel-level identical to the training images. However, they demonstrate significant object-level similarity. 
To quantify such object-level similarity, a commonly used metric is the dot product of embeddings of $\hat{x}_0$ and $n_0$:
\begin{equation}
\vspace{-0.1cm}
\sigma_t = E(\hat{x}_0)^T \cdot E(n_0)
  \label{eq:4}
\vspace{-0.2cm}
\end{equation}
where $E(\cdot)$ represents the embedding obtained via a feature extractor, with Self-supervised Copy Detection (SSCD) \cite{sscd} being the preferred method for identifying object-level memorization \cite{somepalli_2023_cvpr}, $n_0$ represents the nearest neighbor of $x_0$, identified using this metric. 

In later studies \cite{somepalli_2023_neurips, kumari_2023_iccv} that examine memorization issues within the diverse LAION dataset, where numerous object-level memorization instances are identified, SSCD has been established as the standard metric. 
On the other hand, for the CIFAR-10 dataset, the negative nL2 is found to be an efficient measure \cite{carlini_2023_usenix}, attributable to the dataset's smaller size and consistency in image presentation.

\subsection{Causes of Memorization}
\label{sec:mem-sub2}
The main causes of memorization in diffusion models are identified as follows:
1) \textit{Overly specific user prompts} act as a ``key" to the pretrained model's memory, potentially retrieving a specific training image corresponding to this ``key", as observed by \cite{somepalli_2023_neurips}. 
2) \textit{Duplicated training images} are more inclined to be memorized by diffusion models as noted by \cite{carlini_2023_usenix, somepalli_2023_cvpr}, likely due to overfitting.
3) \textit{Duplicated captions across those duplicated images} can exacerbate the memorization issue \cite{somepalli_2023_neurips} by overfitting the text-image pairs to text-conditional diffusion models, turning the caption into a ``key" that consistently retrieves the ``value" of the associated image.
Therefore, when a user employs such repetitive captions or closely related text prompts as the conditioning, the model is prone to generate the corresponding duplicated training image.

%% file: sec/5_methodology_camera_ready.tex
\section{Anti-Memorization Guidance}
\label{sec:method}
\begin{figure}[t]
\centering
\includegraphics[width=1.0\linewidth]{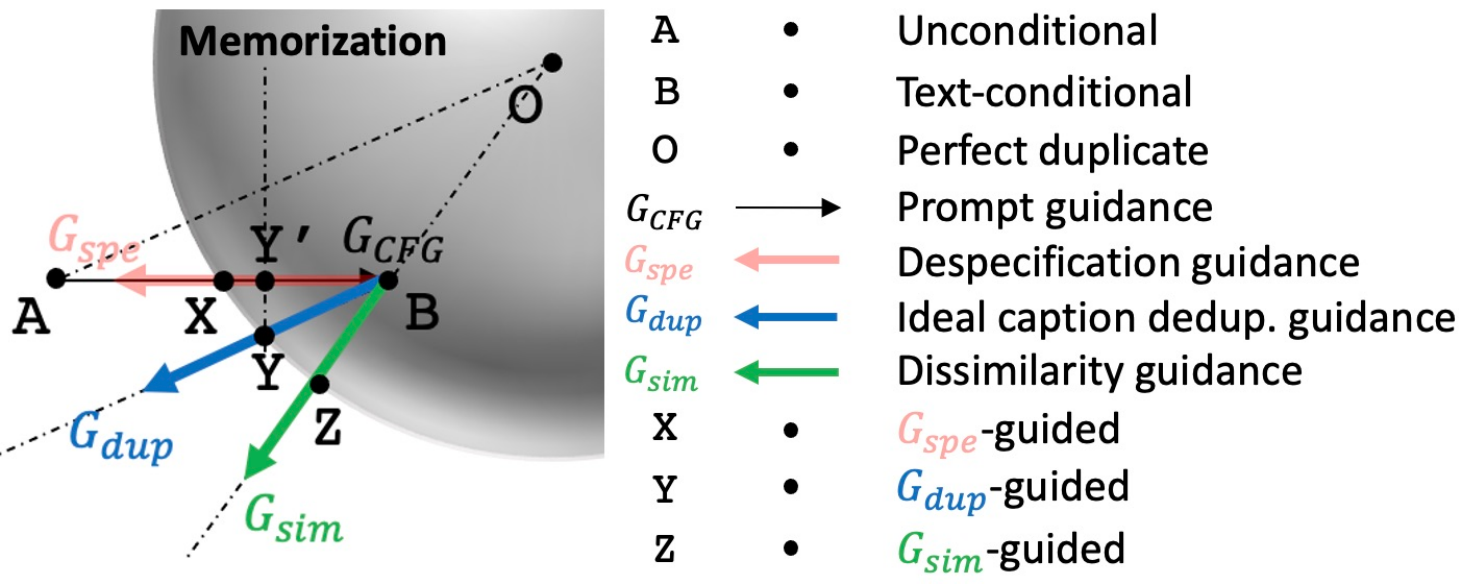}
\caption{Geometric interpretation of different guidance methods and generations. The center \texttt{O} represents a scenario where the generated image is identical to the memorized training image. The distance of any point from \texttt{O} reflects its degree of dissimilarity to the memorized image. The surface of the sphere signifies the threshold that defines the presence of a memorization issue. The arrows represent different types of guidance strategies.}
\label{fig:method}
\vspace{-0.7cm}
\end{figure}
Leveraging insights into the primary causes of memorization, for diffusion models, we present \textit{Anti-Memorization Guidance (AMG)}, a unified framework integrating a comprehensive suite of three distinct guidance strategies, namely, \textit{dissimilarity guidance}, \textit{desspecification guidance}, and \textit{caption deduplication guidance}. 
Each strategy within AMG is meticulously crafted to address and effectively eliminate specific causes of memorization in these models.
As illustrated in \cref{fig:method}, in the original diffusion models employing classifier-free guidance (CFG), the unconditional generation (point \texttt{A}) is linearly guided towards its text-conditional generation (point \texttt{B}), which may falls inside the sphere, indicating that it has become a memorized case.
In the AMG framework, all three guidance methods are capable of steering the generation process away from memorization, represented by moving the generation outside the sphere in the geometric representation.
From an implementation standpoint, each guidance strategy is readily integrable with different types of pretrained diffusion models, such as conditional and unconditional Denoising Diffusion Probabilistic Models (DDPMs) and Latent Diffusion Models (LDMs), without the necessity for additional re-training or fine-tuning. This compatibility extends to different sampling methods, including DDPM sampler and accelerated sampling method such as DDIM \cite{ddim}.
The framework solely requires updating the epsilon prediction $\hat{\epsilon}$ in accordance with each specific guidance strategy, which is then combined with $x_t$ to estimate the previous step representation $x_{t-1}$ in the reverse process of diffusion models:
\begin{equation} 
\hat{\epsilon} \leftarrow \hat{\epsilon} + 1_{\{\sigma_t > \lambda_t\}} \cdot (G_{spe} + G_{dup} + G_{sim})
  \label{eq:amg}
\vspace{-0.6cm}
\end{equation}
\begin{equation}
x_{t-1} \leftarrow \sqrt{\bar{\alpha}_{t-1}} \left( \frac{x_t - \sqrt{1-\bar{\alpha}_t} \hat{\epsilon}}{\sqrt{\bar{\alpha}_t}} \right) + \sqrt{1 - \bar{\alpha}_{t-1}} \hat{\epsilon}
  \label{eq:xt-1}
\end{equation}
To minimize alterations to the original sampling process and thus preserve output utility to the greatest extent, we introduce an indicator function $1_{\{\sigma_t > \lambda_t\}}$ that activates guidance only when the current similarity $\sigma_t$ exceeds a pre-set threshold $\lambda_t$. 
Importantly, we have designed $\lambda_t$ as a dynamic, rather than static, threshold. 
This dynamic nature accounts for the observed fluctuations of $\sigma_t$ throughout the inference process.
For instance, in the early denoising stages, when $t$ is large, the prediction $\hat{x}_0$ is generally less precise than at later stages when $t$ approaches $0$. 
Consequently, $\sigma_t$ tends to be lower at higher $t$ values and increases as $t$ decreases. 
To effectively manage this variation, we adopt a parabolic scheduling for $\lambda_t$ in alignment with the characteristics of the denoising stages. 
As a result, this design of conditional guidance with parabolic scheduling operates as an automatic mechanism to selectively activates guidance only when necessary. Such a configuration enables AMG to optimize the balance between reducing memorization and maintaining high-quality outputs.

Moreover, AMG ensures that the generated images, during inference time, diverge from the memorized training image at either pixel or object level. The type and strength of this divergence can be flexibly tailored based on the user's specific application and objectives.
\subsection{Despecification Guidance}
\label{sec:method-sub1}
As previously discussed, a primary cause of memorization in text-conditional diffusion models is the overly specific nature of user prompts, acting as a ``key" to the pretrained model's memory \cite{somepalli_2023_neurips}.
To reduce caption specificity by instructing the inference process, we firstly employ a \textit{desspecification guidance}. 
Given an noised image or latent-space representation $x_t$ and predicted noise $\hat{\epsilon}$ at time t, we can obtain its prediction of $\hat{x}_0$ using the Diffusion Kernel (\cref{eq:pre3}):
\begin{equation}
\hat{x}_0 = \frac{x_t - \sqrt{1 - \bar{\alpha}_t} \cdot \hat{\epsilon}}{\sqrt{\bar{\alpha}_t}}
  \label{eq:2}
\end{equation}
Then, we search its nearest neighbor $n_0$ in the training set and compute the similarity $\sigma_t$ between $\hat{x}_0$ and $n_0$. 
Depending on the user's goal to prevent pixel-level or object-level memorization, the similarity measure $\sigma_t$ can be computed accordingly using nL2 (\cref{eq:3}) or SSCD (\cref{eq:4}).

This method aligns with the principles of CFG but pursues the inverse goal: to attenuate the original CFG scale to linearly adjust the epsilon prediction to be less aligned with the prompt-conditional prediction:
\begin{equation} 
s_1 =  \max(\min(c_1 \sigma_t, s_0 - 1), 0) 
  \label{eq:5.0}
\vspace{-0.6cm}
\end{equation}
\begin{equation} 
G_{spe} = - s_1 (\epsilon_{\theta}(x_t, y)-\epsilon_{\theta}(x_t)) 
  \label{eq:5}
\end{equation}
where $\epsilon_{\theta}(x_t, y)$ represents the pretrained diffusion model's prediction conditioned on user's text prompt, while $\epsilon_{\theta}(x_t)$ denotes the unconditional prediction.
$s_0$ is the original scale of CFG, $c_1$ is a constant and $c_1 \sigma_t$ defines the guidance scale at step $t$, which is directly proportional to the similarity $\sigma_t$ at step $t$. This enables the algorithm to adaptively adjust the scale of \textit{desspecification guidance} throughout the sampling process, corresponding to the current level of the memorization, as indicated by the value of $\sigma_t$ at any step $t$.
The function $max(\cdot, 0)$ guarantees the guidance scale $-s_1$ to be non-positive, thus diminishing caption specificity, while $min(\cdot)$ function bounds $c_1 \sigma_t$ to not exceed $s_0-1$, safeguarding against excessive low text-image alignment. 

From a geometric perspective as in \cref{fig:method},
Despecification guidance ($G_{spe}$) is capable of steering the generation process away from memorization, start from point \texttt{B}, $G_{spe}$ directs the prediction in the exact opposite direction of the CFG, exiting the sphere at point \texttt{X} on the surface.

\subsection{Caption Deduplication Guidance}
\label{sec:method-sub2}
As outlined in \cref{sec:memorization}, duplicated captions can act as precise ``keys" to retrieve memorized data from the training set, 
so why not turn this to our advantage? By intentionally using them as prompts for pretrained diffusion models to generate predictions that replicates these memorized images, we can then apply classifier-free guidance techniques to steer the generation away from these images:
\begin{equation} 
s_2 =  \max(\min(c_2 \sigma_t, s_0 - s_1 - 1), 0)
  \label{eq:6.0}
\vspace{-0.6cm}
\end{equation}
\begin{equation} 
G_{dup} = - s_2 (\epsilon_{\theta}(x_t, y_N)-\epsilon_{\theta}(x_t)) 
  \label{eq:6}
\end{equation}
where $y_N$ denotes the caption of $n_0$, which is the nearest neighbor of current prediction $\hat{x}_0$ as defined in \cref{eq:2}. 
In case where $n_0$ is a duplicated image prone to memorization and accompanied by a duplicated caption, $y_N$ would correspond to this replicated caption. Consequently, $\epsilon_{\theta}(x_t, y_N)$ reflects the conditional prediction based on the duplicated caption, serving as an ideal antithesis to the prediction we aim to achieve.
The function $max(\cdot, 0)$ again guarantees non-positive guidance scale $-s_2$, directs $\hat{\epsilon}$ away from conditional prediction compared to the unconditional prediction $\epsilon_{\theta}(x_t)$, while $\min(\cdot)$ bounds the total scale of $s_1+c_2 \sigma_t$ to not exceed $s_0-1$ for preserving text-image alignment. 

\begin{figure}[t]
\centering
\includegraphics[width=1.0\linewidth]{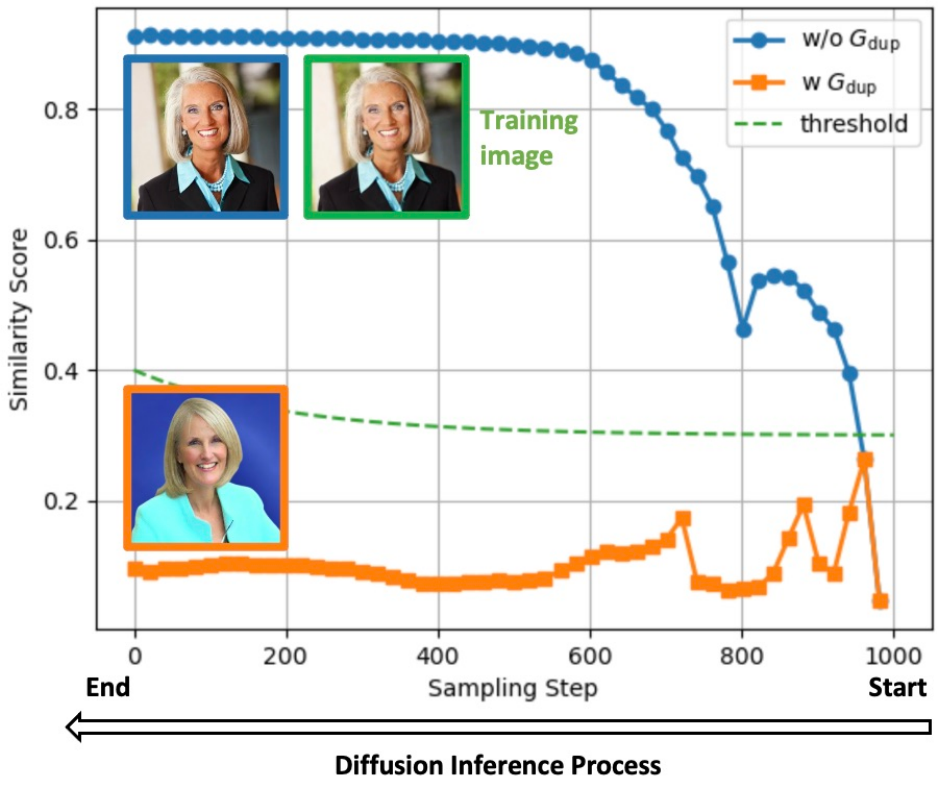}
\caption{Comparison of similarity scores throughout the inference process, with and without the application of $G_{dup}$.}
\label{fig:wo2}
\vspace{-0.5cm}
\end{figure}

From a geometric standpoint as in \cref{fig:method}, caption deduplication guidance ($G_{dup}$) runs parallel to line \texttt{OA}, representing the guidance direction when using perfect text-conditioning that leads to memorization as a negative prompt. This method exits the sphere at point \texttt{Y} on the surface, thereby moving the generation out of the memorization zone.
\cref{fig:wo2} further illustrates the efficiency of $G_{dup}$. It demonstrates that when the similarity score exceeds the dashed parabolic threshold line $\lambda_t$, as defined in \cref{eq:amg}, $G_{dup}$ is activated. This activation prompts $G_{dup}$ to guide the generations in a direction opposite to that of the training image, effectively preventing memorization.

\subsection{Dissimilarity Guidance}
\label{sec:method-sub3}
Distinct from the first two strategies, which linearly adjust the epsilon prediction $\hat{\epsilon}$ along the vector direction between conditional and unconditional predictions, \textit{dissimilarity guidance} identifies another dimension that offer effective guidance to reduce, or even eliminate, memorization in diffusion models.
It extends the discrete class label $y$ in classifier guidance \cref{eq:pre11} to a continuous embedding represented by the similarity score. This approach assures that our generated images are actively directed towards reducing their similarity score, thereby ensuring persistent dissimilarity from their closest counterparts in the training set, as measured by metrics such as nL2 and SSCD.
\begin{equation}
G_{sim} = c_3 \sqrt{1 - \bar{\alpha}_t} \cdot \nabla_{x_t} \sigma_t 
  \label{eq:6}
\end{equation}
where we use the similarity metric $\sigma_t$ instead of log classifier probability $\log p_{\phi}(y|x_t)$ to compute gradient and guide the inference process.
We also invert the sign preceding the guidance term from \cref{eq:pre11} to indicate our new objective: to minimize similarity as opposed to maximizing it.
An additional scaling factor $c_3$ is also employed, which functions as a hyperparameter to control the intensity of the guidance, thereby managing the privacy-utility trade-off and tailoring user's specific goals. 
A larger $c_3$ permits greater deviation of the generated data from the nearest training image, but at the expense of reduced quality and text alignment.

Geometrically as in \cref{fig:method}, dissimilarity guidance ($G_{sim}$) steers the generation direction directly away from point \texttt{O}, which represents the perfectly memorized case (\ie, the training data), eventually exit at point \texttt{Z} on the sphere.

\subsection{Unpacking AMG's Threefold Guidance}
\label{sec:method-sub4}
We present a detailed analysis of the indispensable role and synergies of each of the three guidance methods in AMG for achieving the optimal balance between privacy and utility. 

\textbf{Impact on quality and text-alignment.}
Similar to CFG, our $G_{spe}$ method linearly combines the unconditional prediction $\epsilon_{\theta}(x_t)$ and user-prompt based conditional predictions $\epsilon_{\theta}(x_t, y)$.
Altering the guidance scale modifies the weights of these components, but as both are derived from pretrained diffusion model's high-quality outputs, overall output quality remains largely consistent. 
However, text alignment decreases with lower weights on $\epsilon_{\theta}(x_t)$, necessitating a minimum weight of one to preserve text alignment.
Similarly, for $G_{dup}$, assuming using a neighbor's caption $y_N$ leads to an exact replication of training image, the output maintains high quality, thus its linear combination with $\epsilon_{\theta}(x_t)$ also yields quality on par with the pretrained model.
Geometrically, results within the \texttt{ABY} plane, formed by lines \texttt{AB} and \texttt{BY}, maintain quality, but those closer to \texttt{A} (along line \texttt{AB}) show reduced text alignment.
$G_{sim}$, however, does not align with the \texttt{ABY} plane, so its scale affects the quality and must be minimally set to preserve quality.

\textbf{Importance of dissimilarity guidance ($G_{sim}$).} 
$G_{sim}$, as shown in \cref{fig:method}, is vital despite its possible quality impact. The necessity stems from the fact that $G_{spe}$ and $G_{dup}$'s combined scale, capped at 
$s_0-1$, cannot assure moving the generation outside the sphere, potentially leaving it within the \texttt{BXY} sector. $G_{sim}$, on the other hand, reliably ensures the generation is guided out of the sphere, effectively addressing memorization.

\textbf{Importance of caption deduplication guidance ($G_{dup}$).} 
Text-conditioning heightens specificity, thus reducing diversity in generations. In severe cases, such as when point \texttt{B} near center \texttt{O}, $G_{sim}$ would need a high scale to prevent memorization without $G_{spe}$ and $G_{dup}$, risking artifacts. $G_{spe}$ and $G_{dup}$ mitigate this by lowering the needed scale for $G_{sim}$, thus preserving output quality.
Comparing $G_{spe}$ and $G_{dup}$, both maintain quality within the \texttt{ABY} plane, but $G_{dup}$ involves less text alignment sacrifice due to the shorter linear projection of \texttt{BY} on the \texttt{AB} line (\ie, \texttt{BY'} $<$ \texttt{BX}), thus more beneficial than $G_{spe}$.

\textbf{Importance of despecification guidance ($G_{spe}$).} 
The inclusion of $G_{spe}$, despite $G_{dup}$'s apparent advantages, is justified by $G_{dup}$'s practical limitations as it idealizes deduplication guidance, requiring access to perfect ``keys" for pretrained model memories, approximated here by captions of memorized training images. Such ideal ``keys" are uncommon, and caption-based approximations may not always be as precise as prompt-based methods in $G_{spe}$, particularly when memorized images' captions are not duplicated in the dataset. Consequently, our approach in AMG integrates both $G_{spe}$ and $G_{dup}$ to emulate these ideal ``keys" for effective negative guidance.

\textbf{Unconditional and class-conditional generations.}
Our research reveals that in the absence of text-conditioning, memorization in diffusion models is reduced but not eliminated, aligning with our prior findings. This significantly lessens memorization, leaving image duplication as the only remaining issue.
In such cases, $G_{spe}$ is highly effective in eliminating memorization due to two factors:
1) \textit{Early Detection}: During initial stages of reverse diffusion, our conditional guidance with a parabolic schedule efficiently identifies potential replication. 2) \textit{Increased Diversity}: 
Without specific text-conditioning, the generation process yields greater diversity.
This enables $G_{dis}$ to effectively steer generations away from memorized modes to un-memorized ones in the initial stages, ensuring they don't revert. Once memorization ceases to be detected, further guidance application is discontinued. 
As a result, this method predominantly impacts the coarse structure, with guidance typically applied only during the early stages of the denoising process. 
The finer details in later stages remain intact, thus ensuring the overall quality of the output is preserved.

%% file: sec/6_evaluation_camera_ready.tex
\section{Experiments}
\label{sec:results}
\subsection{Experimental Setup}
\label{sec:setup}
\label{sec:comparisons}
\begin{figure}[t]
\centering
\includegraphics[width=1.0\linewidth]{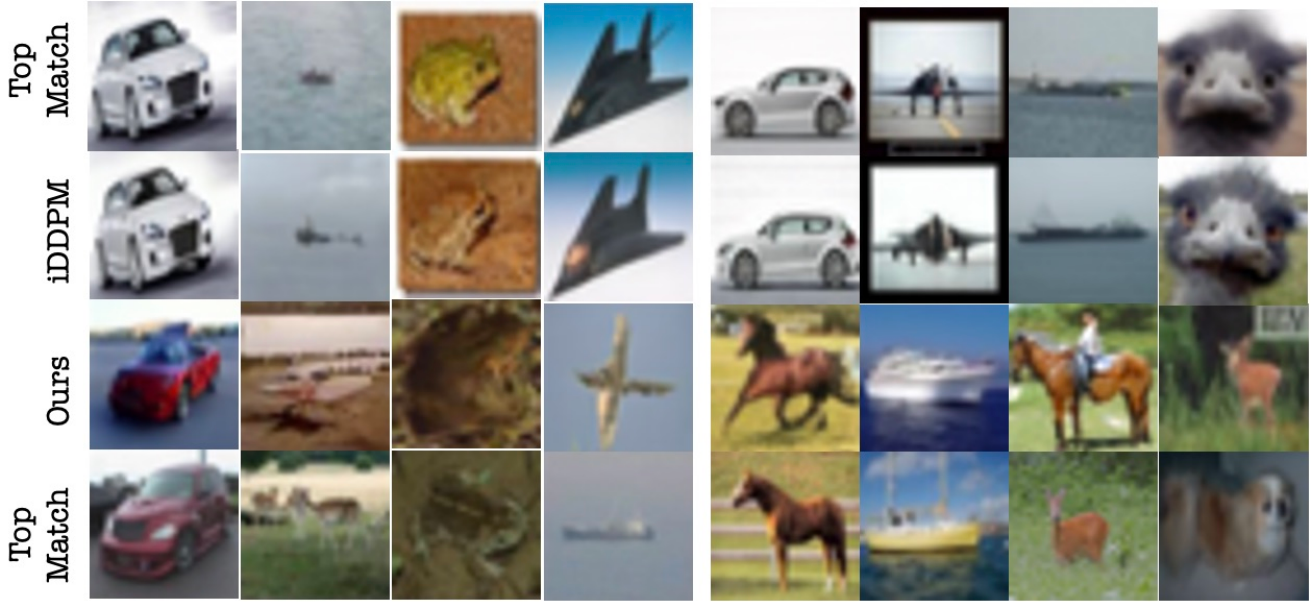}
\caption{Applying AMG to iDDPM on CIFAR-10. Left: Class-conditional generation. Right: Unconditional generation.}
\label{fig:cifar-250}
\vspace{-0.4cm}
\end{figure}
\textbf{Scope}. 
In the realm of pretrained diffusion models, studies \cite{carlini_2023_usenix, somepalli_2023_neurips, somepalli_2023_cvpr} have identified memorization in unconditional DDPMs on CIFAR-10 and text-conditional Stable Diffusion on LAION datasets. While \cite{somepalli_2023_cvpr} found no memorization in latent diffusion models using ImageNet, our analysis of DDPMs on ImageNet \cite{imagenet} and LSUN Bedroom \cite{lsun} also showed no memorization cases.
Beyond these scenarios, we explored class-conditional generation and identified memorization cases on CIFAR-10. Our findings confirm the successful elimination of memorization in all tested scenarios, highlighting the wide applicability of our approach.

\textbf{Evaluation metrics.} 
To assess text-conditional generations (\cref{table:laion}), we employ three metric types: memorization, quality, and text-alignment. Memorization metrics include the 95th percentile \cite{somepalli_2023_neurips} of similarity scores of all generated images, determined using pixel-level nL2 norm (\cref{eq:3}) or object-level SSCD embedding similarity (\cref{eq:4}). 
We contend that relying solely on this metric might be misleading, especially if the distribution of the generated data exhibits a heavy upper tail beyond the 95th percentile. In such scenarios, the similarity score could be significantly understated.
We propose to additionally examine the maximum similarity score within the distribution, to gauge the worst-case scenario regarding memorization. 
Additionally, the proportion of images exceeding certain similarity thresholds, thus flagged as memorized, is a key metric \cite{carlini_2023_usenix}. Notably, thresholds vary; for CIFAR-10, an nL2 below $1.4$ normally indicates pixel-level memorization, while for LAION, an SSCD above $0.5$ suggests object-level memorization following the convention of previous work \cite{somepalli_2023_cvpr, somepalli_2023_neurips, kumari_2023_iccv}. We use FID to measure the fidelity and diversity of generated images, and CLIP score to measure the generated images' alignment with the input text prompts.

\textbf{Implementational details.} 
Experimental results depend on variables such as text prompts, number of generated images, training image scope (e.g., LAION-10k to LAION-5B), choice of diffusion model, and sampling steps. 
Since baseline methods often employ different settings, we have reimplemented these baselines to ensure a fair and comparable evaluation.
In our text-conditional generation experiments on the LAION5B dataset, we utilized \cite{beaumont-2022-clip-retrieval}'s system for replicates identification. They used a CLIP embedding-based index for LAION5B, with an efficient retrieval system for identifying k-nearest neighbors. Our approach differed in seeking the singular nearest neighbor based on SSCD embedding similarity. To approximate this, we first identified 1,000 images with the lowest CLIP embedding similarities, then computed SSCD similarities to find the highest match, using it for memorization metrics. 
Stable Diffusion v1.4 with a DDIM sampler and 50 sampling steps was our model of choice.
For unconditional and class-conditional generations on CIFAR-10, which comprises 50,000 images, we calculated SSCD similarities for each generated image with the entire training set. OpenAI's iDDPM~\cite{iddpm_2021_icml} with a DDPM sampler and 250 sampling steps was used. 
Further details are in the supplementary material.
\vspace{-0.3cm}
\subsection{Comparison with Baselines}
\vspace{-0.cm}
\begin{table}[t] 
\begin{center}
\small
\scriptsize
\begin{tabular}{M{0.15\linewidth} | M{0.08\linewidth} M{0.08\linewidth} M{0.08\linewidth} M{0.08\linewidth} | M{0.08\linewidth} | M{0.10\linewidth} }
\hline
\hline
 & \multicolumn{4}{c|}{Memorization Metrics by SSCD $\downarrow$} & & \\ 
 & Top5\% & Top1 & \%$>$0.5 & \%$>$0.4 & FID $\downarrow$ & CLIP $\uparrow$ \\
\hline
\textcolor{black}{SD~\cite{sd_2022_cvpr}} & 0.91 & 0.93 & 44.85 & 59.23 & 106.41 & \bf{28.04} \\
\hline
\textcolor{black}{Ablation~\cite{kumari_2023_iccv}} & - & - & 0.30* & - & - & - \\
\hline
\textcolor{black}{GNI~\cite{somepalli_2023_neurips}} & 0.91 & 0.94 & 42.75 & 58.18 & 97.81 & 27.79 \\
\hline
\textcolor{black}{RT~\cite{somepalli_2023_neurips}} & 0.61 & 0.84 & 15.07 & 26.75 & 101.69 & 22.63 \\
\hline
\textcolor{black}{CWR~\cite{somepalli_2023_neurips}} & 0.79 & 0.85 & 26.45 & 40.93 & \bf{96.25} & 25.96 \\
\hline
\textcolor{black}{RNA~\cite{somepalli_2023_neurips}} & 0.75 & 0.82 & 17.78 & 29.05 & 99.68 & 23.37 \\
\hline
\textcolor{black}{Ours(Main)} & 0.41 & 0.47 & \bf{0.00} & 7.07 & 99.12 & 26.98 \\
\hline
\textcolor{black}{Ours(Strong)} & \bf{0.34} & \bf{0.39} & \bf{0.00} & \bf{0.00} & 100.45 & 26.72 \\
\hline
\hline
\end{tabular}
\end{center}
\vspace{-0.4cm}
\caption{Comparisons on text-conditional generation of LAION5B based on SSCD similarity. AMG successfully eliminates memorization with minimal impact on quality and text-alignment.} 
\vspace{-0.3cm}
\label{table:laion}
\end{table}
\begin{table}[t] 
\vspace{-0.cm}
\begin{center}
\small
\scriptsize
\begin{tabular}{M{0.15\linewidth} | M{0.08\linewidth} M{0.08\linewidth} M{0.08\linewidth} M{0.08\linewidth} | M{0.08\linewidth} }
\hline
\hline
 & \multicolumn{4}{c|}{Memorization Metrics by nL2} & \\ 
 & Top5\%$\uparrow$ & Top1$\uparrow$ & \%$<$1.4$\downarrow$ & \%$<$1.6$\downarrow$ & FID$\downarrow$ \\
\hline
\textcolor{black}{iDDPM~\cite{iddpm_2021_icml}} & 1.58 & 0.51 & 0.93 & 5.78 & 7.44 \\
\hline
\textcolor{black}{Ours(Main)} & 1.61 & 1.47 & \bf{0.00} & 4.34 & 7.25 \\
\hline
\textcolor{black}{Ours(Strong)} & \bf{1.71} & \bf{1.68} & \bf{0.00} & \bf{0.00} & \bf{6.98} \\
\hline
\hline
\end{tabular}
\end{center}
\vspace{-0.4cm}
\caption{Comparisons on unconditional generation of CIFAR-10 based on nL2 similarity. AMG effectively eliminates memorization without affecting image quality.} 
\vspace{-0.3cm}
\label{table:cifar-unc-sscd}
\end{table}
\begin{table}[t] 
\begin{center}
\small
\scriptsize
\begin{tabular}{M{0.15\linewidth} | M{0.08\linewidth} M{0.08\linewidth} M{0.08\linewidth} M{0.08\linewidth} | M{0.08\linewidth} }
\hline
\hline
 & \multicolumn{4}{c|}{Memorization Metrics by nL2} & \\ 
 & Top5\%$\uparrow$ & Top1$\uparrow$ & \%$<$1.4$\downarrow$ & \%$<$1.6$\downarrow$ & FID$\downarrow$ \\
\hline
\textcolor{black}{iDDPM~\cite{iddpm_2021_icml}} & 1.53 & 0.51 & 1.53 & 9.77 & 11.81 \\
\hline
\textcolor{black}{Ours(Main)} & 1.56 & 1.46 & \bf{0.00} & 8.70 & 11.54 \\
\hline
\textcolor{black}{Ours(Strong)} & \bf{1.71} & \bf{1.68} & \bf{0.00} & \bf{0.00} & \bf{11.44} \\
\hline
\hline
\end{tabular}
\end{center}
\vspace{-0.4cm}
\caption{Comparisons on class-conditional generation of CIFAR-10 based on nL2 similarity. AMG effectively eliminates memorization without affecting image quality.} 
\vspace{-0.3cm}
\label{table:cifar-c-sscd}
\end{table}
\textbf{Text-conditional generations on LAION.} 
\Cref{table:laion} shows that AMG outperforms all baselines in memorization metrics by a huge margin, eliminating all memorization cases defined by a similarity score over $0.5$. It also shows a minimal loss in text-alignment, evidenced by having the second-highest CLIP score among mitigation strategies, thus maintaining strong alignment with user intentions. Notably, the strategy with the highest CLIP score, GNI, performs poorly in memorization metrics, closely resembling the original Stable Diffusion without mitigation. AMG matches baselines in FID, indicating comparable quality. Overall, AMG leads to memorization-free generations and stands out in balancing quality and utility.
AMG's flexibility allows users to adjust guidance strength based on their definition of memorization. While the standard threshold is $0.5$, increasing AMG's guidance scale (\ie, the strong version of AMG) effectively prevents memorization even at a $0.4$ threshold, at a minimal extra cost of quality and text-alignment.

\textbf{Unconditional and class-conditional generations on CIFAR-10.}
\cref{table:cifar-unc-sscd} and \cref{table:cifar-c-sscd} illustrate AMG's effectiveness in transitioning from text-conditional to class-conditional or unconditional generation tasks, and from preventing object-level to pixel-level memorization. 
AMG consistently outperforms, even when compared to \cite{carlini_2023_usenix}, who reported a 23\% reduction in memorization by retraining diffusion models on a deduplicated CIFAR-10 dataset. 
AMG ensures memorization-free outputs and even slightly exceeds the original diffusion model's quality, as reflected in FID scores, likely due to the increased diversity of its generated images compared to the original model's replicated outputs.
This success is attributed to two main factors:
1) AMG's early memorization detection during reverse sampling, utilizing a conditional guidance with a parabolic schedule;
2) The absence of text-conditioning eliminates key memorization causes, like specific user prompts and caption duplication, enhancing output diversity. Solely using \textit{dissimilarity guidance} in AMG can be very effective in preventing memorization whilst preserving output quality, since it only alters the coarse structure of images in early stages of reverse sampling when potential memorization is detected. Guidance ceases once memorization is no longer detected, thus preserving sample quality.
\subsection{Ablation Studies}
\label{sec:ablation}
\begin{table}[t] 
\begin{center}
\small
\scriptsize
\begin{tabular}{M{0.20\linewidth} | M{0.10\linewidth} M{0.10\linewidth} | M{0.10\linewidth} | M{0.10\linewidth} }
\hline
\hline
 & \multicolumn{2}{c|}{Mem. by SSCD $\downarrow$} & & \\ 
 & Top5\% & \%$>$0.5 & FID $\downarrow$ & CLIP $\uparrow$ \\
\hline
\textcolor{black}{Baseline~\cite{sd_2022_cvpr}} & 0.9133 & 44.85 & 106.41 & \bf{28.04} \\
\hline
\textcolor{black}{$G_{sim}+G_{spe}$} & 0.4072 & \bf{0.00} & \color{gray}{119.13} & \color{gray}{26.67} \\
\hline
\textcolor{black}{$G_{sim}+G_{dup}$} & 0.4073 & \bf{0.00} & \color{gray}{120.48} & \color{gray}{26.17} \\
\hline
\textcolor{black}{$G_{spe}+G_{dup}$} & \color{gray}{0.7396} & \color{gray}{31.62} & \bf{87.10} & 27.18 \\
\hline
\textcolor{black}{Full} & \bf{0.4066} & \bf{0.00} & 99.12 & 26.98 \\
\hline
\end{tabular}
\end{center}
\vspace{-0.4cm}
\caption{Ablation studies on text-conditional generation based on SSCD. Grey-colored font denotes areas of sacrifice.} 
\vspace{-0.3cm}
\label{table:ablation}
\end{table}

\begin{table}[t] 
\begin{center}
\small
\scriptsize
\begin{tabular}{M{0.30\linewidth} | M{0.10\linewidth} M{0.10\linewidth} | M{0.10\linewidth} }
\hline
\hline
 & \multicolumn{2}{c|}{Mem. by nL2} & \\ 
 & Top5\%$\uparrow$ & \%$<$1.4$\downarrow$ & FID $\downarrow$ \\
\hline
\textcolor{black}{Baseline~\cite{iddpm_2021_icml}} & 1.58 & 0.93 & 7.44 \\
\hline
\textcolor{black}{w/o conditional guidance} & 1.49 & \bf{0.00} & \color{gray}{257.27} \\
\hline
\textcolor{black}{w constant schedule} & 1.59 & \color{gray}{0.04} & 7.44 \\
\hline
\textcolor{black}{Full} & \bf{1.61} & \bf{0.00} & \bf{7.25} \\
\hline
\end{tabular}
\end{center}
\vspace{-0.4cm}
\caption{Ablation studies on unconditional generation based on nL2. Grey-colored font denotes areas of sacrifice.} 
\vspace{-0.3cm}
\label{table:ablation-cifar}
\end{table}
\Cref{table:ablation} underlines the crucial role of AMG's tripartite guidance in optimizing the privacy-utility trade-off. Key observations include:
1) All AMG versions notably enhance memorization metrics, particularly those incorporating $G_{sim}$, which eradicate memorization entirely with proper guidance scale tuning. This underscores $G_{sim}$'s theoretical guarantee against memorization, albeit with slight impacts on quality and text-alignment. Ablation of $G_{sim}$ improves FID and CLIP scores but results in 31.62\% of generated images being marked as memorized. Thus, $G_{sim}$ inclusion significantly boosts privacy with minimal utility loss.
2) Comparisons between the full AMG version and variants lacking either $G_{dup}$ or $G_{spe}$ demonstrate that while achieving similar privacy levels, the ablated versions yield inferior FID and CLIP scores. This confirms the importance of both $G_{dup}$ and $G_{spe}$ in the guidance ensemble.

\Cref{table:ablation-cifar} highlights the efficacy of our conditional guidance with parabolic scheduling. Applying guidance indiscriminately during sampling, rather than selectively based on potential memorization, guarantees elimination of memorization but degrades output quality, evidenced by significantly higher FID scores. This is due to excessive alteration of both coarse structures (early sampling stages) and finer details (later stages).
The parabolic schedule aligns with denoising stages: initially, predictions are highly noised and dissimilar to training images, becoming more accurate and revealing potential memorization cases with higher similarity as denoising progresses. This schedule enables early detection and effective resolution of memorization issues. A constant schedule would fail to provide this early detection, leading to 0.04\% of generations being memorized, which could be eliminated by increasing the guidance scale but at the cost of quality. Therefore, our conditional guidance strategy enhances the privacy-utility trade-off by negating the need for this additional quality compromise.

%% file: sec/7_conclusions_camera_ready.tex
\vspace{-0.2cm}
\section{Conclusion}
\label{sec:conclusion}
\vspace{-0.cm}
We introduce \textit{AMG}, a unified framework featuring three specialized guidance strategies, each addressing a specific cause of memorization in diffusion models. Theoretical analysis and empirical ablation studies confirm the essential role of each strategy in achieving an optimal privacy-utility trade-off. AMG's strategic guidance scheduling and innovative automatic detection enable conditional application, further refining this balance. Our experiments demonstrate that AMG reliably generates images during inference that are distinct from memorized training images, maintaining high quality and text-alignment. 
Furthermore, AMG offers the flexibility to adapt to various user requirements by allowing customization in the type of memorization prevented (pixel-level or object-level) through adjustments in the similarity metrics employed in its guidance. Additionally, it provides options for guidance intensity (main or strong version) by adjusting the guidance scale, catering to a wide range of applications and user preferences.

\section{Acknowledgement}
\label{sec:acknowledgement}
This work was supported in part by the Australian Research Council under Projects DP210101859 and FT230100549.
\vspace{-0.8cm}

%% file: sec/X_suppl.tex
\clearpage
\maketitlesupplementary
\appendix
\label{sec:appendix}
\textbf{Outline}. 
In \cref{sec:conditional_guidance}, we analyze additional insights into the synergistic use of Anti-Memorization Guidance (AMG) with our deliberately designed conditional guidance strategy and parabolic scheduling.
\cref{sec:kdeplots} introduces KDE plots as an alternative evaluative method for memorization, showcasing the distribution of the top 1 similarity score across generated images, complementing the main paper's quantitative results.
\cref{sec:adaptability} demonstrates AMG's adaptability in switching samplers or similarity metrics within its guidance to meet specific user needs, maintaining effectiveness in memorization eradication.
\cref{sec:imp} delves into the implementation details.
Finally, \cref{sec:qualitative} offers additional qualitative results.
\section{Additional Analysis}
\label{sec:conditional_guidance}
The power of AMG can be amplified when paired with our deliberately designed conditional guidance strategy that incorporates a parabolic scheduling threshold for determining the activation of AMG at each inference step.

\begin{figure}[t]
\centering
\includegraphics[width=1.0\linewidth]{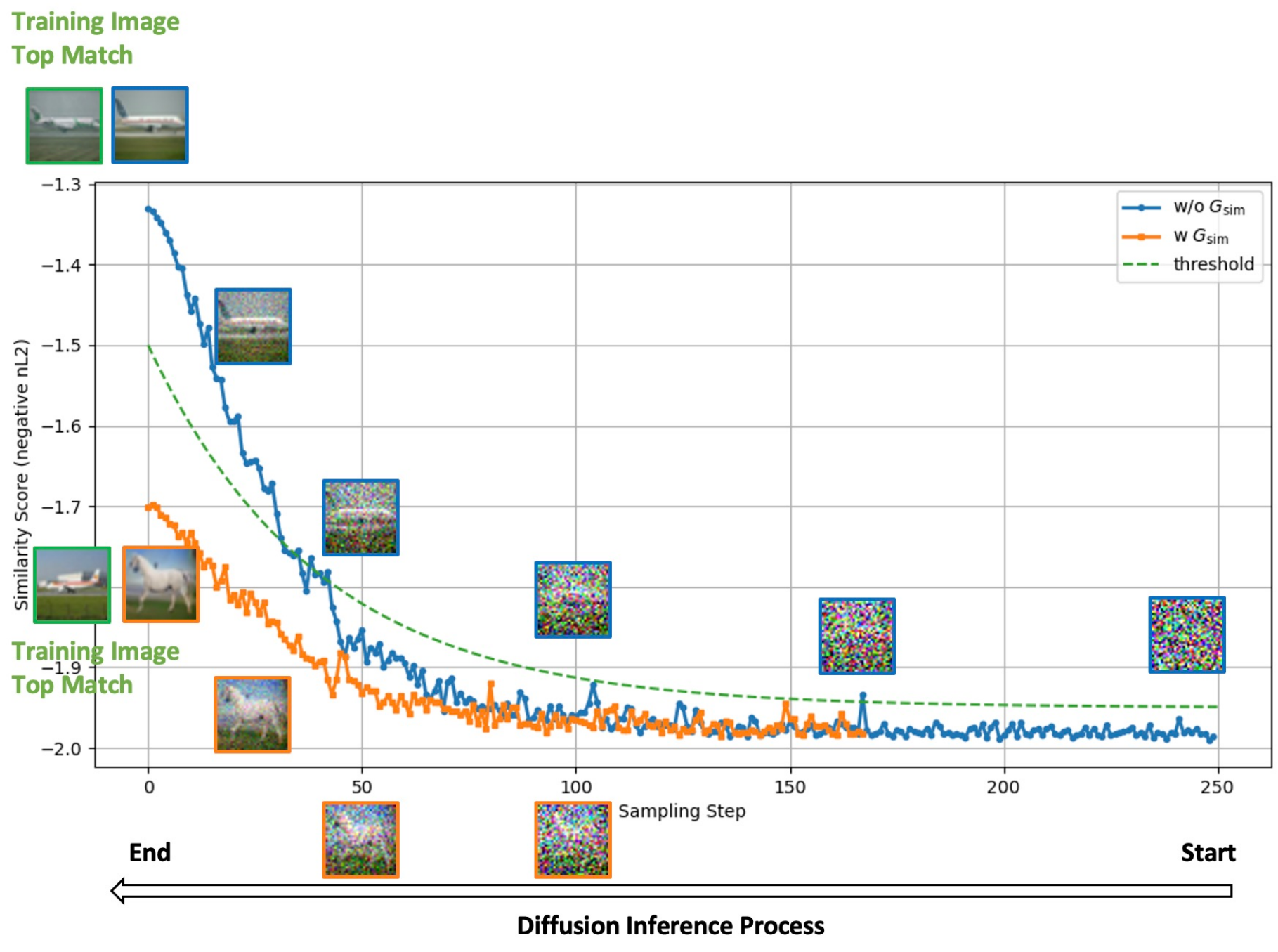}
\caption{Example illustrating the detection of potential memorization at an \textbf{early} stage of reverse diffusion process, enabling AMG to influence the coarser structures in generated outputs.}
\label{fig:abl_170}
\end{figure}

\begin{figure}[t]
\centering
\includegraphics[width=1.0\linewidth]{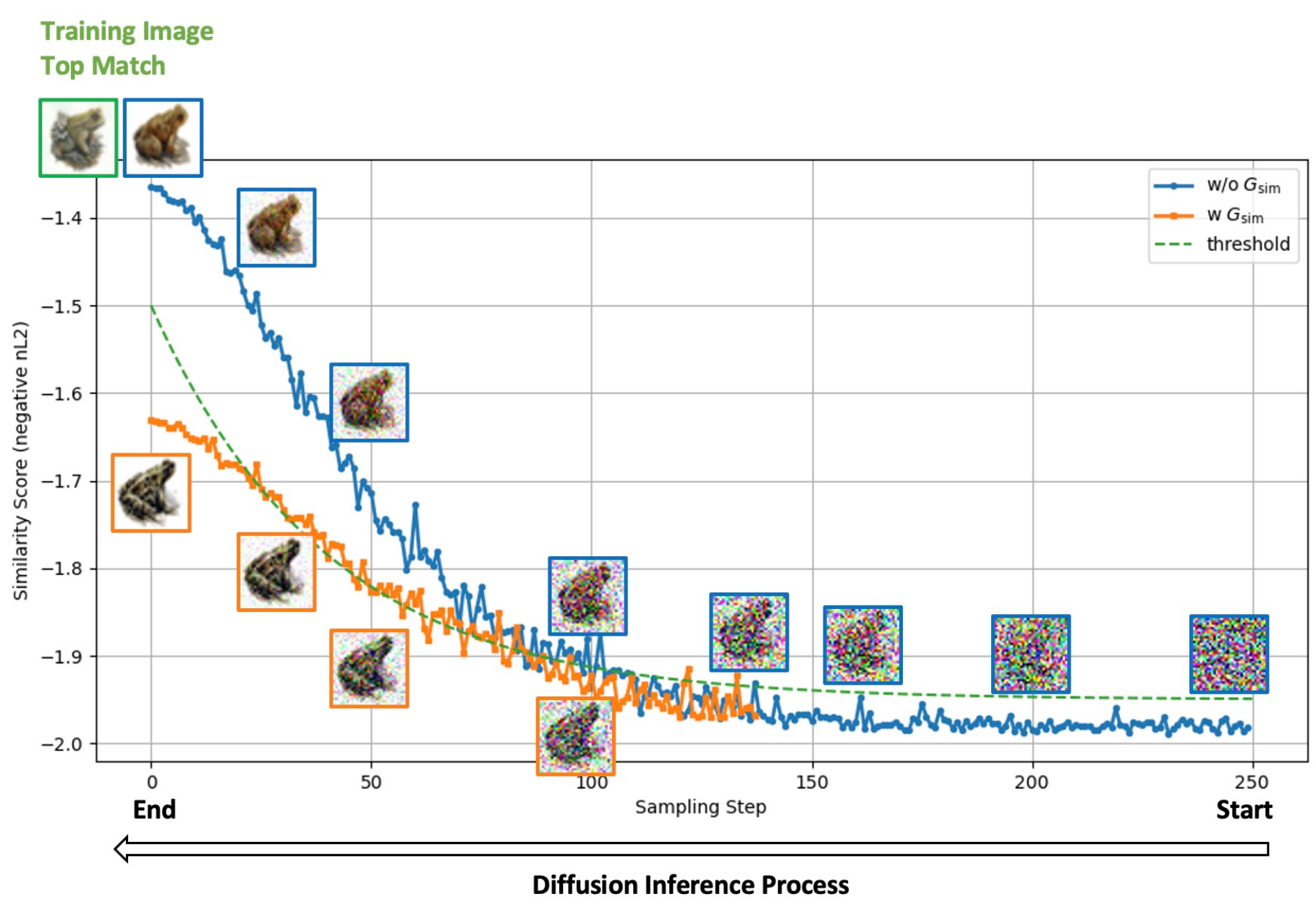}
\caption{Example illustrating the early detection of potential memorization at an \textbf{intermediate} stage of reverse diffusion process, enabling AMG to influence the generation's structures that are between coarse and fine detail.}
\label{fig:abl_140}
\end{figure}

\begin{figure}[t]
\centering
\includegraphics[width=1.0\linewidth]{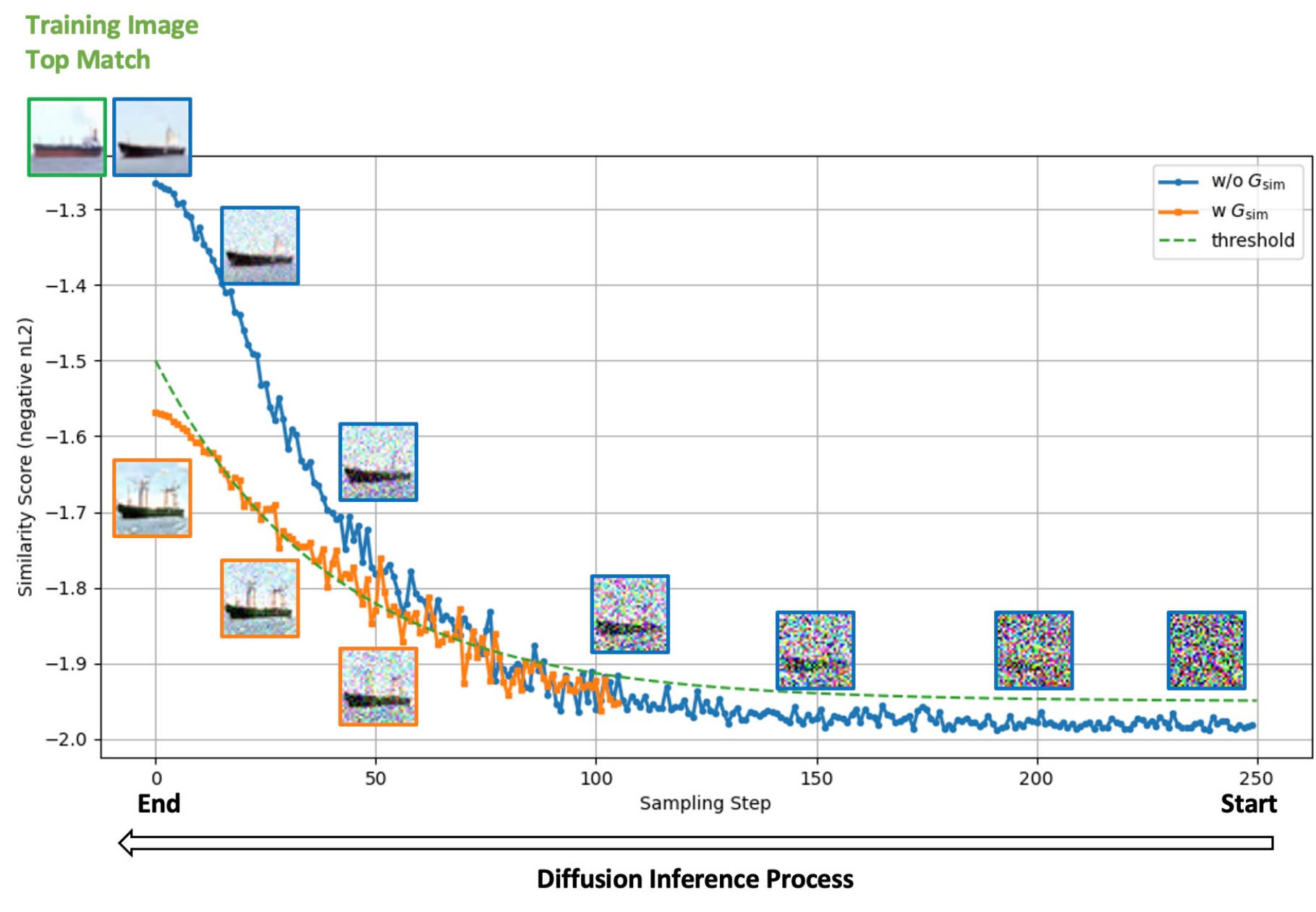}
\caption{Another example illustrating the early detection of potential memorization at an \textbf{intermediate} stage of reverse diffusion process, enabling AMG to influence the generation's structures that are between coarse and fine detail.}
\label{fig:abl_100}
\end{figure}

\begin{figure}[t]
\centering
\includegraphics[width=1.0\linewidth]{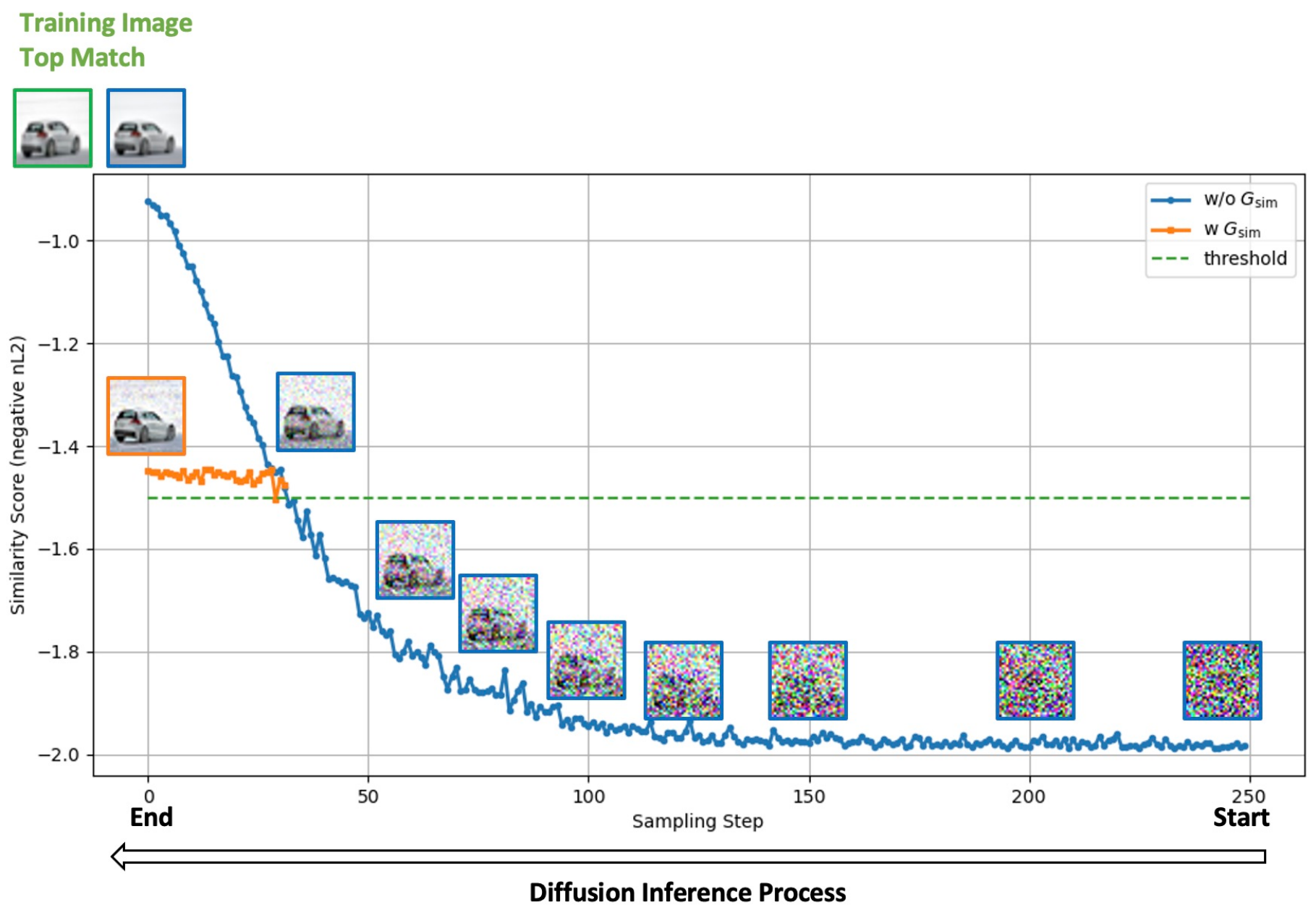}
\caption{An \textbf{ablated counterexample} highlighting the risk of incomplete memorization prevention and the consequence of altering only the \textbf{finest} details and \textbf{injecting noises to image background} to lower the similarity score when employing a constant guidance schedule, as opposed to the more effective parabolic scheduling in our conditional guidance approach.}
\label{fig:abl_con_2}
\end{figure}

\begin{figure}[t]
\centering
\includegraphics[width=1.0\linewidth]{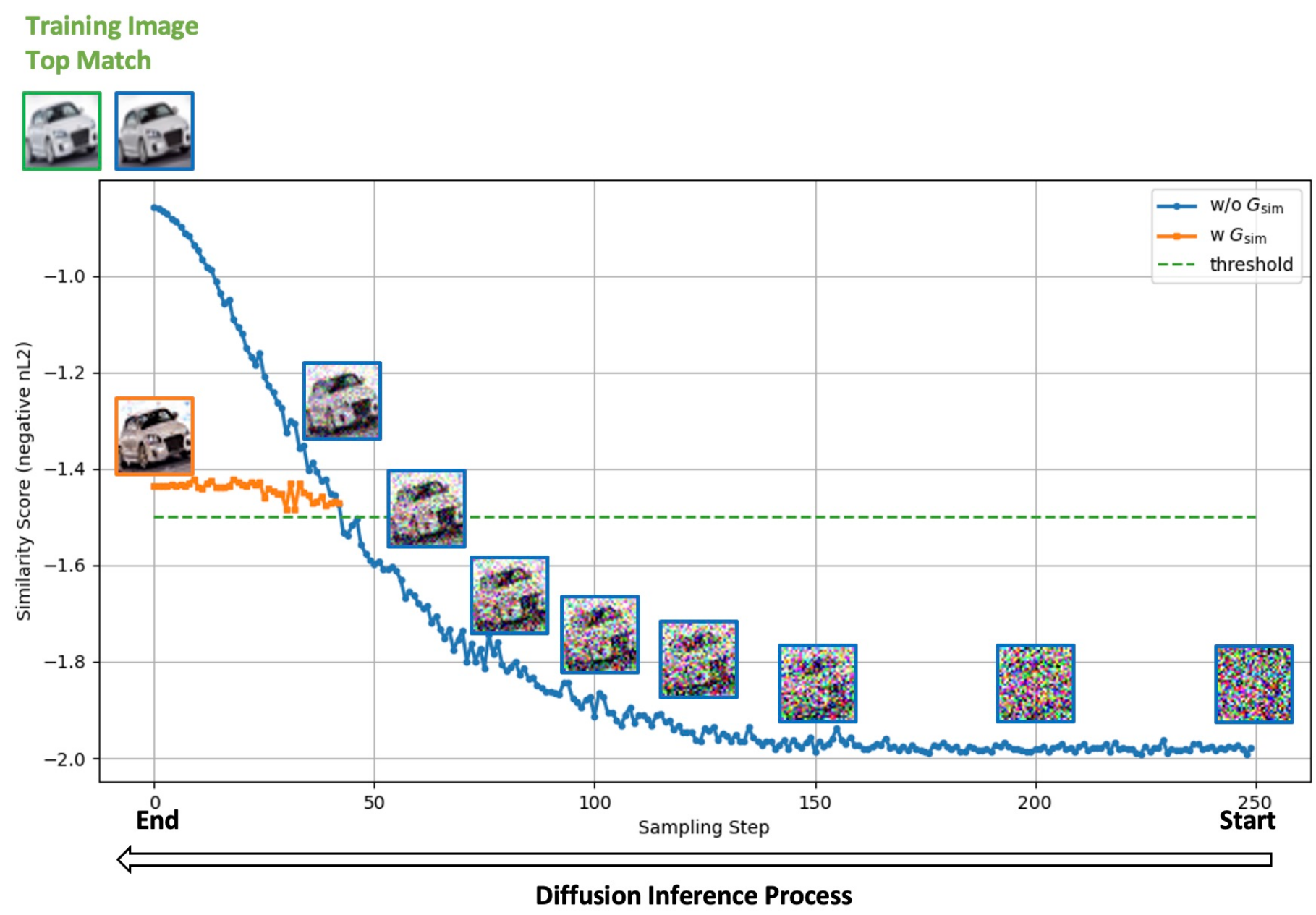}
\caption{Another \textbf{ablated counterexample} highlighting the risk of incomplete memorization prevention and the consequence of altering only the \textbf{finest} details and \textbf{injecting noises to image background} to lower the similarity score when employing a constant guidance schedule, as opposed to the more effective parabolic scheduling in our conditional guidance approach.}
\label{fig:abl_con_1}
\end{figure}

Firstly, the efficacy of $G_{sim}$, our dissimilarity guidance component, is clear in \cref{fig:abl_170} to \cref{fig:abl_con_1}, with guided versions consistently registering lower similarity scores compared to their ablated counterparts, indicating successful deviation from memorized training images:
\begin{equation}
G_{sim} = c_3 \sqrt{1 - \bar{\alpha}_t} \cdot \nabla_{x_t} \sigma_t 
  \label{eq:sup1}
\end{equation}
where $c_3$ further enables users to control the guidance intensity, thus balancing privacy and utility. This is also capable of providing a guarantee of memorization-free generations by simply setting a large $c_3$ at the cost of output quality. \textbf{Pairing $G_{sim}$ with our conditional guidance strategy can optimize such trade-off}.

Specifically, our conditional guidance uses a parabolic schedule (\cref{eq:sup_parabolic}) that is closely aligned with the characteristics of the denoising stages as can be observed from the blue lines in \cref{fig:abl_170} to \cref{fig:abl_con_1}, where early-stage predictions have lower similarity scores, which is then increased exponentially in the following mathematical form:
\begin{equation}
\lambda_t = a + (b-a) e^{-ct}
  \label{eq:sup_parabolic}
\end{equation}

This distinctive pattern of similarity scores during the denoising stages can be attributed to the initial noised and imprecise predictions, which register exceedingly low similarity scores according to nL2 (\cref{eq:sup2}). As $t$ decreases, the denoising process yields more distinct predictions $\hat{x}_t$, resulting in elevated similarity scores that can potentially reveal cases of memorization.
\begin{equation}
\sigma_t = - \frac{\ell_2(\hat{x}_t,n_t)}{\alpha \cdot \frac{1}{k} \sum_{z_t \in S_{\hat{x}_t}} \ell_2(\hat{x}_t,z_t)}
  \label{eq:sup2}
\end{equation}

Our parabolic schedule is tailored to this trend, serving as the threshold for AMG's activation. This design facilitates the early detection and intervention of potential memorization instances, as demonstrated in \cref{fig:abl_170}, \cref{fig:abl_140}, and \cref{fig:abl_100}.
Specifically, in \cref{eq:sup_parabolic}, the parameter $a$ represents the asymptotic threshold value that the parabolic schedule approaches as $t$ increases towards infinity, which we have set to $-1.95$. The parameter $b$ denotes the value of the parabolic schedule at $t=0$, and we have set this to $-1.5$, intentionally lower than the threshold defining memorization, which is $-1.4$. $c$ controls the shape of the parabolic schedule, which we set to $-0.025$ to produce the green dashed lines in \cref{fig:abl_170} to \cref{fig:abl_con_1}.

\cref{fig:abl_con_2} and \cref{fig:abl_con_1}, serving as two counterexamples, underscores the importance of timely intervention. Utilizing a constant guidance schedule results in late memorization detection, necessitating larger late-stage adjustments. These adjustments often involve last-ditch efforts like adding noise to the image background, noticeable upon close inspection in \cref{fig:abl_con_2} and \cref{fig:abl_con_1}, to reduce the similarity score. Such measures can diminish image quality and might still fail to effectively prevent memorization.

By identifying and intervening memorization early, AMG primarily affects the coarser structures, preserving the finer details and overall aesthetic of the generated images. This approach avoids resorting to last-minute measures to lower the similarity score at the cost of quality. Thus, this optimizes the privacy-utility trade-off, results in more visually appealing outputs that are distinct from the training data, as demonstrated in \cref{fig:abl_170} to \cref{fig:abl_100}.

\section{Alternative Evaluative Method}
\label{sec:kdeplots}
In the main paper, we first evaluate memorization using two key quantitative metrics in accordance with the established standards in the literature: (1) the 95th percentile of the top 1 similarity scores of all generated images, as per \cite{somepalli_2023_neurips}, and (2) the proportion of images exceeding certain similarity thresholds, indicating memorization, following \cite{carlini_2023_usenix}. Additionally, we assess the maximum similarity score to understand the worst-case scenario. Relying solely on the 95th percentile might not fully capture the distribution, particularly if there's a significant upper tail beyond this percentile.

\begin{figure}[t]
\centering
\includegraphics[width=1.0\linewidth]{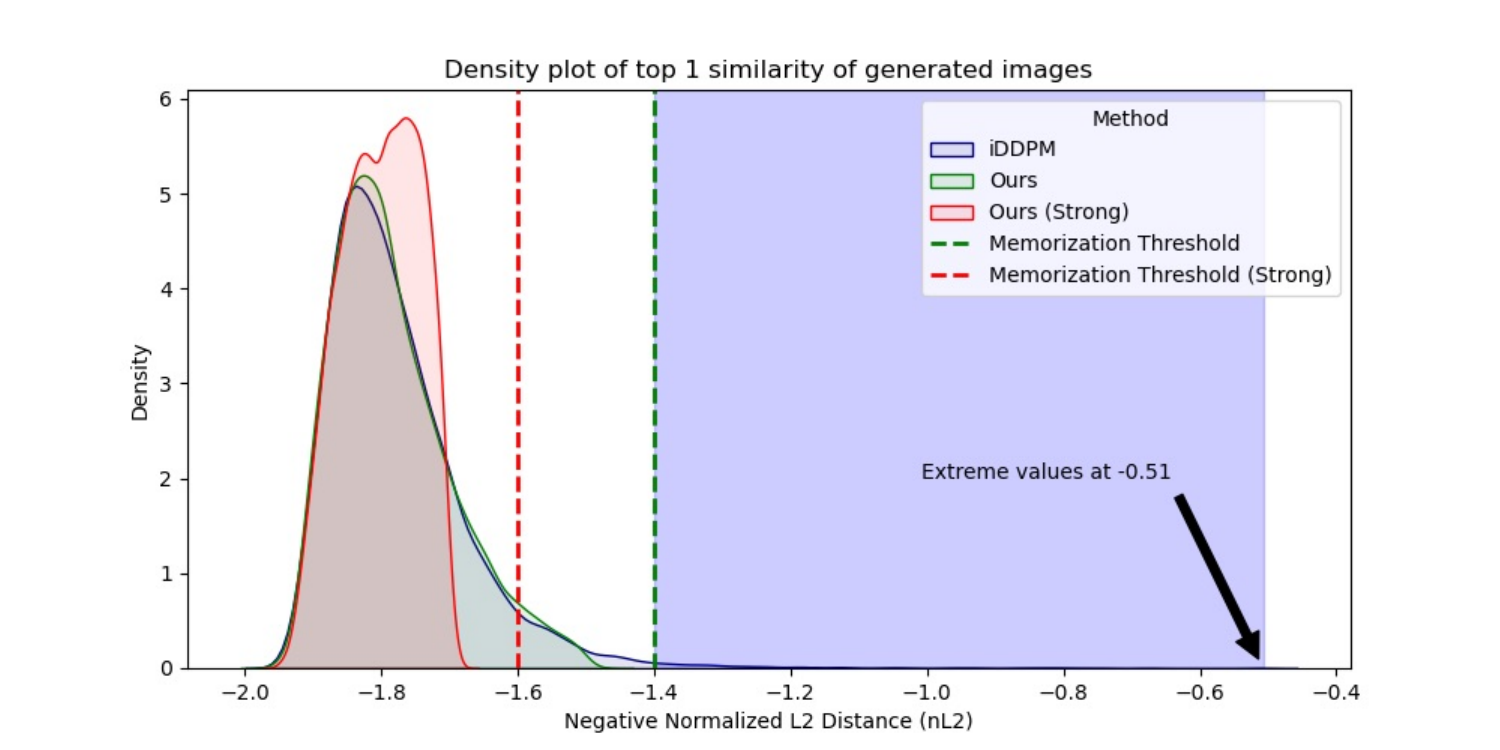}
\caption{KDE plot depicting the top 1 similarity scores for \textbf{unconditional CIFAR-10 generation}. AMG effectively shifts all outputs to the left, maintaining similarity scores below the established memorization threshold.}
\label{fig:kdeplot_uc}
\end{figure}

\begin{figure}[t]
\centering
\includegraphics[width=1.0\linewidth]{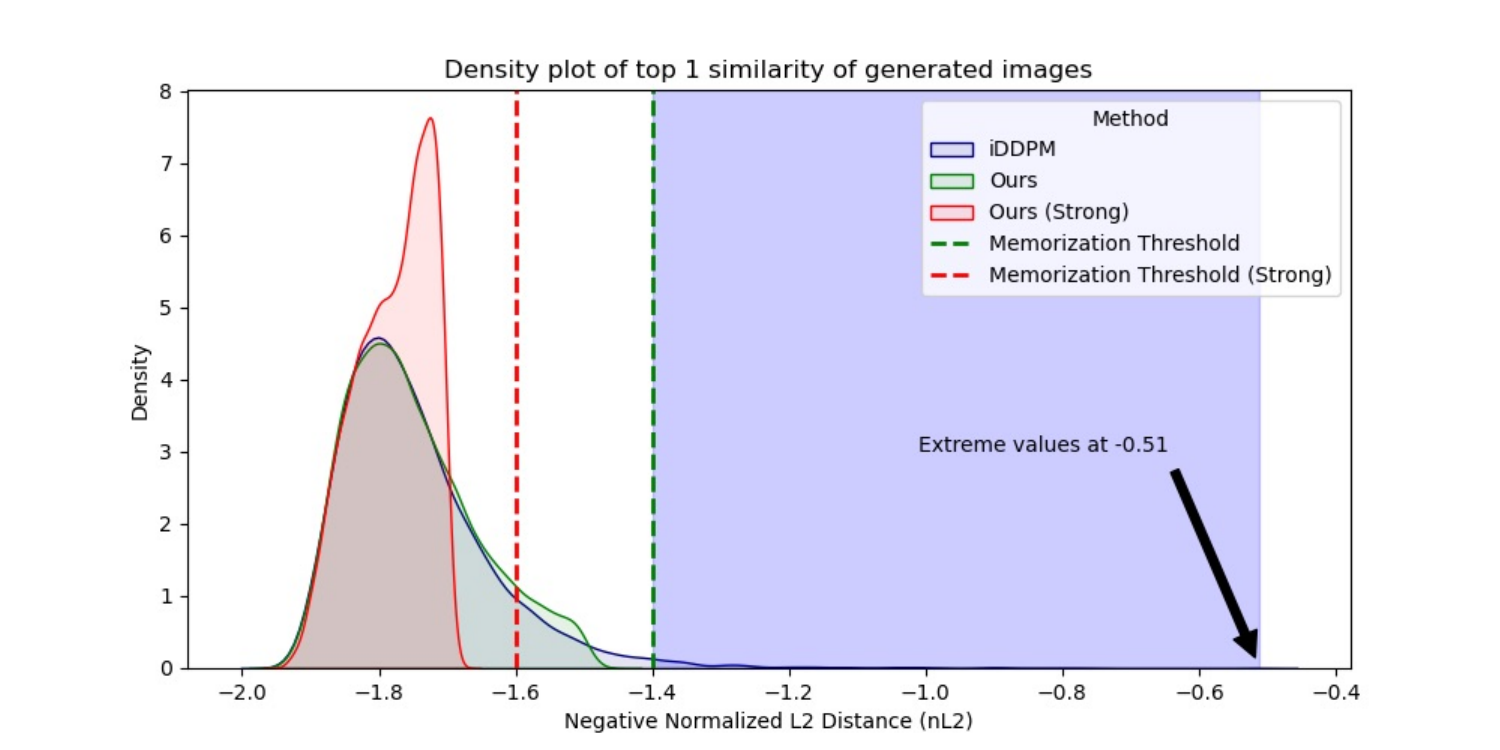}
\caption{KDE plot depicting the top 1 similarity scores for \textbf{class-conditional CIFAR-10 generation}. AMG effectively shifts all outputs to the left, maintaining similarity scores below the established memorization threshold.}
\label{fig:kdeplot_c}
\end{figure}

\begin{figure}[t]
\centering
\includegraphics[width=1.0\linewidth]{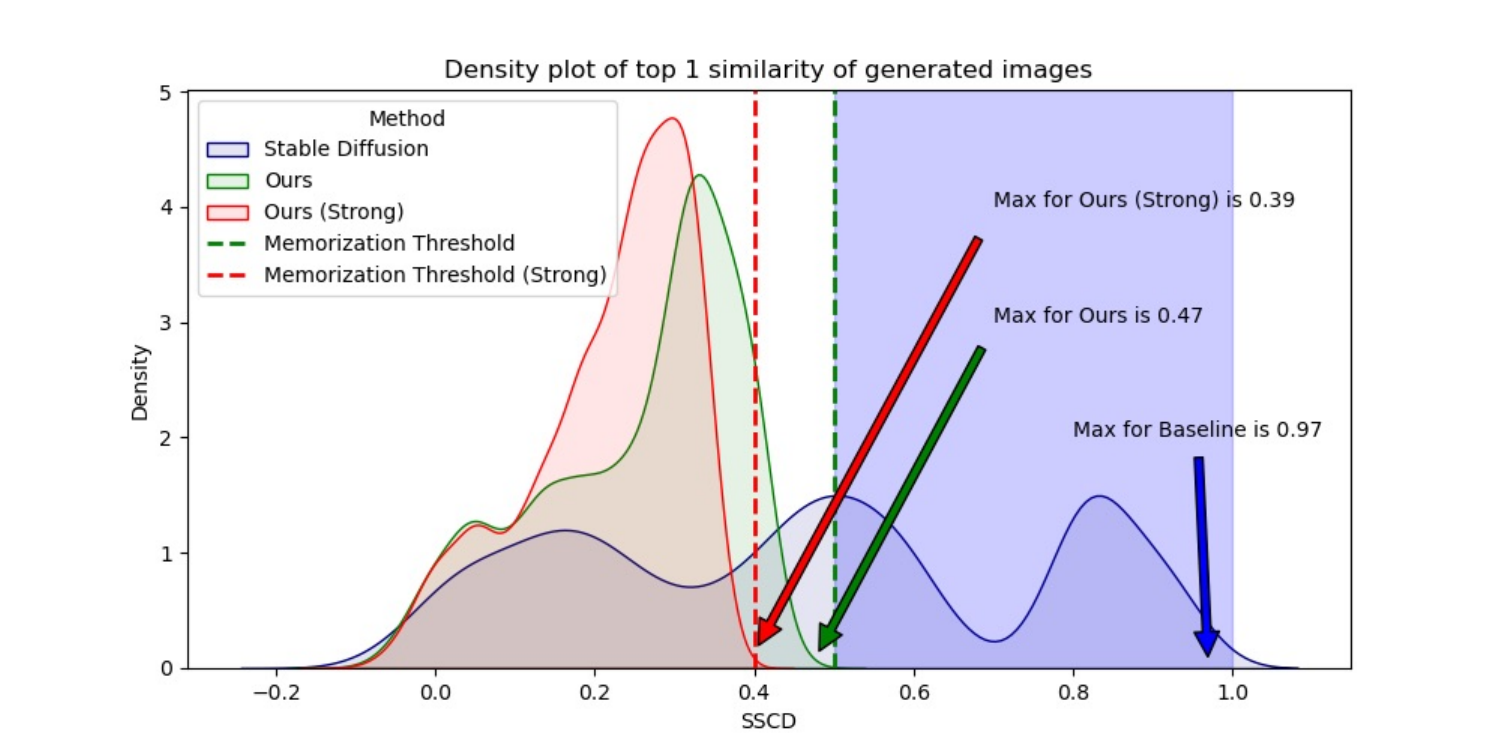}
\caption{KDE plot depicting the top 1 similarity scores for \textbf{text-conditional LAION generation}. AMG effectively shifts all outputs to the left, maintaining similarity scores below the established memorization threshold.}
\label{fig:kdeplot_laion}
\end{figure}

However, these metrics still don't fully represent the distribution of the top 1 similarity scores. To address this, we introduce qualitative KDE (Kernel Density Estimation) plots as a complementary qualitative evaluation method for memorization, showcased in \cref{fig:kdeplot_uc}, \cref{fig:kdeplot_c}, and \cref{fig:kdeplot_laion}, providing a more comprehensive view of the data distribution.

Text-conditional image generation presents a greater memorization challenge, as depicted in \cref{fig:kdeplot_laion}, where a larger portion of the blue curves (representing the top 1 similarity scores of pretrained models) cross the memorization threshold compared to unconditional and class-conditional generations in \cref{fig:kdeplot_uc} and \cref{fig:kdeplot_c}. 
Irrespective of the generation task, AMG reliably shifts these distributions leftward, reducing similarity to training data. The main AMG version ensures all top 1 similarity scores fall below the conventional thresholds ($-1.40$ for CIFAR-10 and $0.50$ for LAION), and the strong AMG version meets even stricter thresholds ($-1.60$ for CIFAR-10 and $0.40$ for LAION), affirming its efficacy.

\section{AMG's Wide Adaptability}
\label{sec:adaptability}
This section illustrates AMG's flexibility as a unified framework adaptable to different similarity measures for guidance signals and evaluation, as well as to different sampling methods including DDPM and accelerated techniques like DDIM \cite{ddim}, while retaining its efficacy.

\subsection{Switching Similarity Measures}
\label{sec:cifar10_sscd}
Self-supervised Copy Detection (SSCD) has emerged as the preferred method for detecting memorization in the LAION dataset \cite{somepalli_2023_cvpr, somepalli_2023_neurips, kumari_2023_iccv}, outperforming other metrics like the normalized L2 distance (nL2) and CLIP. 
Although nL2 has been the metric of choice for CIFAR-10 in prior studies \cite{carlini_2023_usenix}, no work has verified its superiority to other measures such as SSCD.
Thus, to complement, we also adopt SSCD as the guiding and evaluation metric within AMG, as an alternative to nL2. This change allows SSCD to influence guidance activation and scale, and also to direct updates in the prediction process as described in \cref{eq:sup1}. 
For CIFAR-10, we find a threshold of $0.50$ for top 1 SSCD similarity does not accurately indicate memorization, often resulting in false positives. Therefore, we have adjusted the threshold, considering instances with top 1 SSCD similarity scores above $0.90$ as true memorization cases to improve the test's precision.
\cref{table:cifar-unc-sscd} shows that AMG retains its effectiveness in eliminating memorization even when transitioning to SSCD, confirming the framework's flexibility and robustness.

\begin{table}[t] 
\vspace{-0.cm}
\begin{center}
\small
\scriptsize
\begin{tabular}{M{0.15\linewidth} | M{0.10\linewidth} M{0.10\linewidth} M{0.10\linewidth} | M{0.10\linewidth} }
\hline
\hline
 & \multicolumn{3}{c|}{Memorization Metrics by SSCD $\downarrow$} & \\ 
 & Top5\% & Top1 & \%$>$0.90 & FID$\downarrow$ \\
\hline
\textcolor{black}{iDDPM~\cite{iddpm_2021_icml}} & 0.83 & 0.96 & 0.42 & \bf{7.44} \\
\hline
\textcolor{black}{Ours} & \bf{0.79} & \bf{0.85} & \bf{0.00} & 7.58 \\
\hline
\hline
\end{tabular}
\end{center}
\vspace{-0.5cm}
\caption{Comparisons on unconditional generation of CIFAR-10 based on SSCD similarity. AMG effectively eliminates memorization without affecting image quality.} 
\vspace{-0.3cm}
\label{table:cifar-unc-sscd}
\end{table}

\subsection{Switching Samplers}
\label{sec:ddpm_sampler}
In the main paper, we detail the implementation of our Anti-Memorization Guidance (AMG) framework utilizing the DDIM sampler. For tasks with lower computational demands such as unconditional and class-conditional generation on the CIFAR-10 dataset, employing the DDPM sampler is a practical alternative. With the DDPM sampler, the dissimilarity guidance $G_{sim}$ in AMG framework simplifies to the following expressions:
\begin{equation}
G_{sim} = c \cdot \nabla_{x_t} \sigma_t 
  \label{eq:sup_Gsim_new}
\vspace{-0.6cm}
\end{equation}
\begin{equation}
x_{t-1} \leftarrow \text{sample from } \mathcal{N}(\mu - 1_{\{\sigma_t > \lambda_t\}} \cdot \Sigma G_{sim}, \Sigma)
  \label{eq:sup_xt-1_new}
\end{equation}
where $\mu$ and $\Sigma$ represent the mean and variance of the model's predicted distributions, respectively. These adjustments maintain the integrity of AMG's guidance while accommodating the operational characteristics of the DDPM sampler.

\section{Implementational Details}
\label{sec:imp}
This section provides additional implementational details, please refer to the code for even more comprehensive details, which is also included in the supplementary material.

\textbf{Applying AMG on Latent Diffusion Models (LDMs)}.
As discussed in the main paper, to compute similarity scores, we need to obtain the model's prediction of $\hat{x}_0$ using the following Diffusion Kernel:
\begin{equation}
\hat{x}_0 = \frac{x_t - \sqrt{1 - \bar{\alpha}_t} \cdot \hat{\epsilon}}{\sqrt{\bar{\alpha}_t}}
  \label{eq:2}
\end{equation}
This presumes the representation $x_t$ is in the pixel space across the diffusion steps from $t=T$ to $0$, so that we can then search for its nearest neighbor $n_0$ in the training set, which is also in the pixel space.
However, in the context of LDMs, $\hat{x}_0$ would be in latent space, thus it necessitates an additional conversion step using the decoder $\mathcal{D}$ from LDM's pre-trained autoencoder to obtain the pixel-space representation: $\hat{x}_0 \leftarrow \mathcal{D}(\hat{x}_0)$, before searching for its nearest training image in this updated pixel-space representation.

\textbf{Scope of memorization eradication}.
Our research ambitiously targets the most practically significant scope of eliminating memorization in diffusion models, specifically ensuring the generation of images with a similarity score below a commonly accepted threshold (\eg, SSCD $<0.50$) across the extensive LAION5B dataset. 

Initially daunting due to the sheer volume of comparisons required, the task was made feasible by utilizing the official clip-retrieval tool \cite{beaumont-2022-clip-retrieval} provided by the LAION5B dataset creator, which includes a pre-trained k-nearest neighbor index for the entire LAION5B dataset. Specifically, its ClipClient allows remote querying of a clip-retrieval backend via python for very efficient retrieval of nearest neighbors based on the dot product of CLIP embeddings.

A challenge arises from the discrepancy between the CLIP-based retrieval provided by ClipClient and the SSCD embedding we use for object-level similarity. To bridge this gap, for each generated image, we retrieve the closest 1000 neighbors according to CLIP embeddings and then compute SSCD similarities to pinpoint the closest match. While this introduces a small risk of missing the global SSCD minimum, it's manageable by adjusting the search breadth at additional or less computational costs. 

Furthermore, our approach allows for targeted scrutiny of specific images, where the user can directly supply URLs of the images of interest, circumventing broader searches to focus on preventing memorization of chosen images.
Implementationally, our method necessitates only the image URLs as additional input. The algorithm automates the process thereafter, loading the URLs as images, computing their SSCD embeddings, and ensuring these specified images are included in the SSCD similarity computation, regardless of their retrieval status via ClipClient.

This personalized approach can drastically cut computational demands and refine the search to user-defined priorities. It also encompasses the narrower scope adopted in baseline methods like those in \cite{somepalli_2023_neurips}, treating them as a subset scenario where our methodology is adapted to their set of 10,000 selected LAION5B images, sidelining the broader ClipClient search.

\section{Additional Qualitative Results}
\label{sec:qualitative}
\begin{figure*}[!ht]
  \centering
  \includegraphics[width=\textwidth]{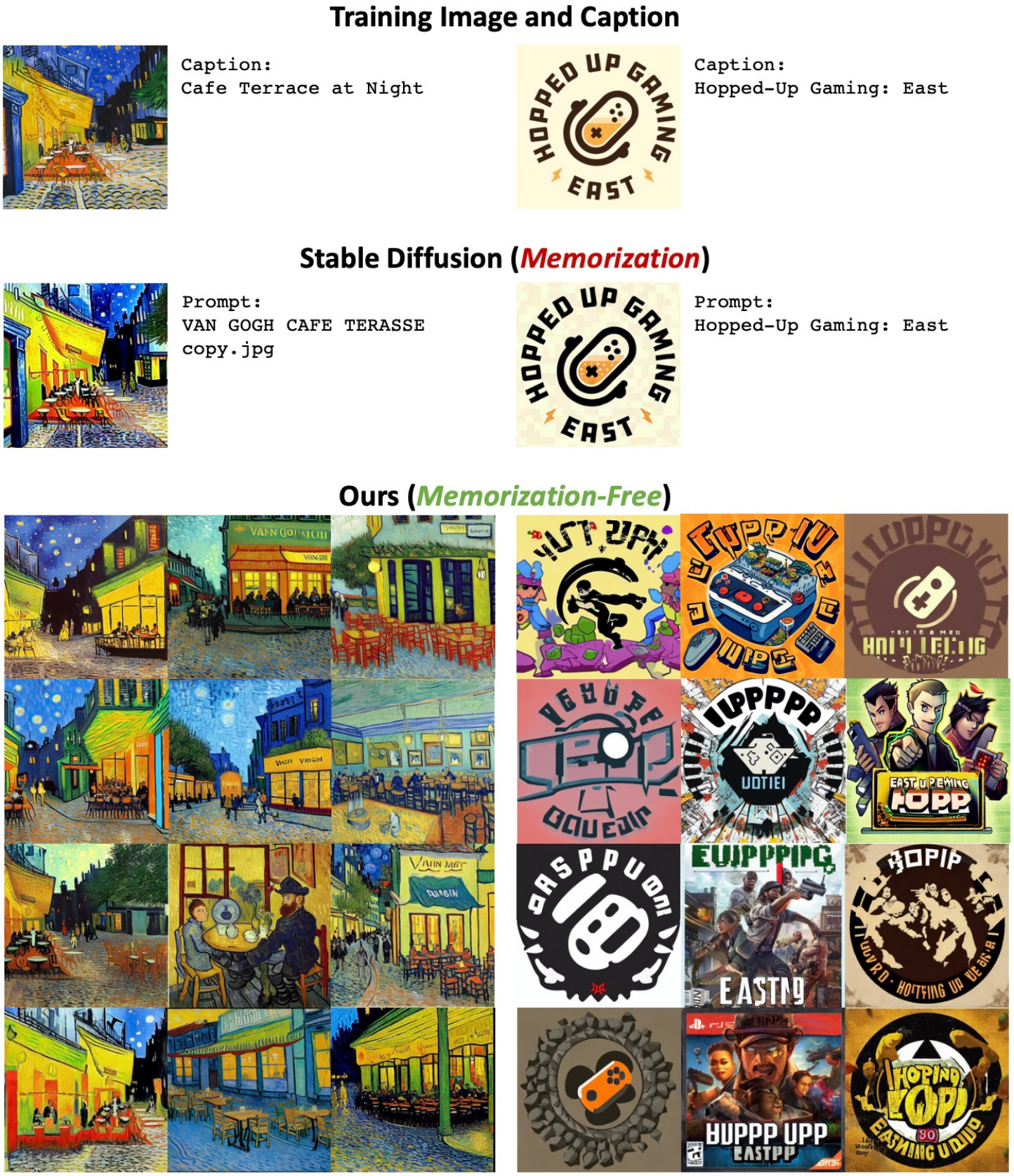}
  \captionsetup{font=large} 
  \caption{Additional qualitative comparisons showcase AMG's effectiveness in guiding pretrained diffusion models to produce memorization-free outputs.}
  \label{fig:qualitative8_6}
\end{figure*}

\begin{figure*}[!ht]
  \centering
  \includegraphics[width=\textwidth]{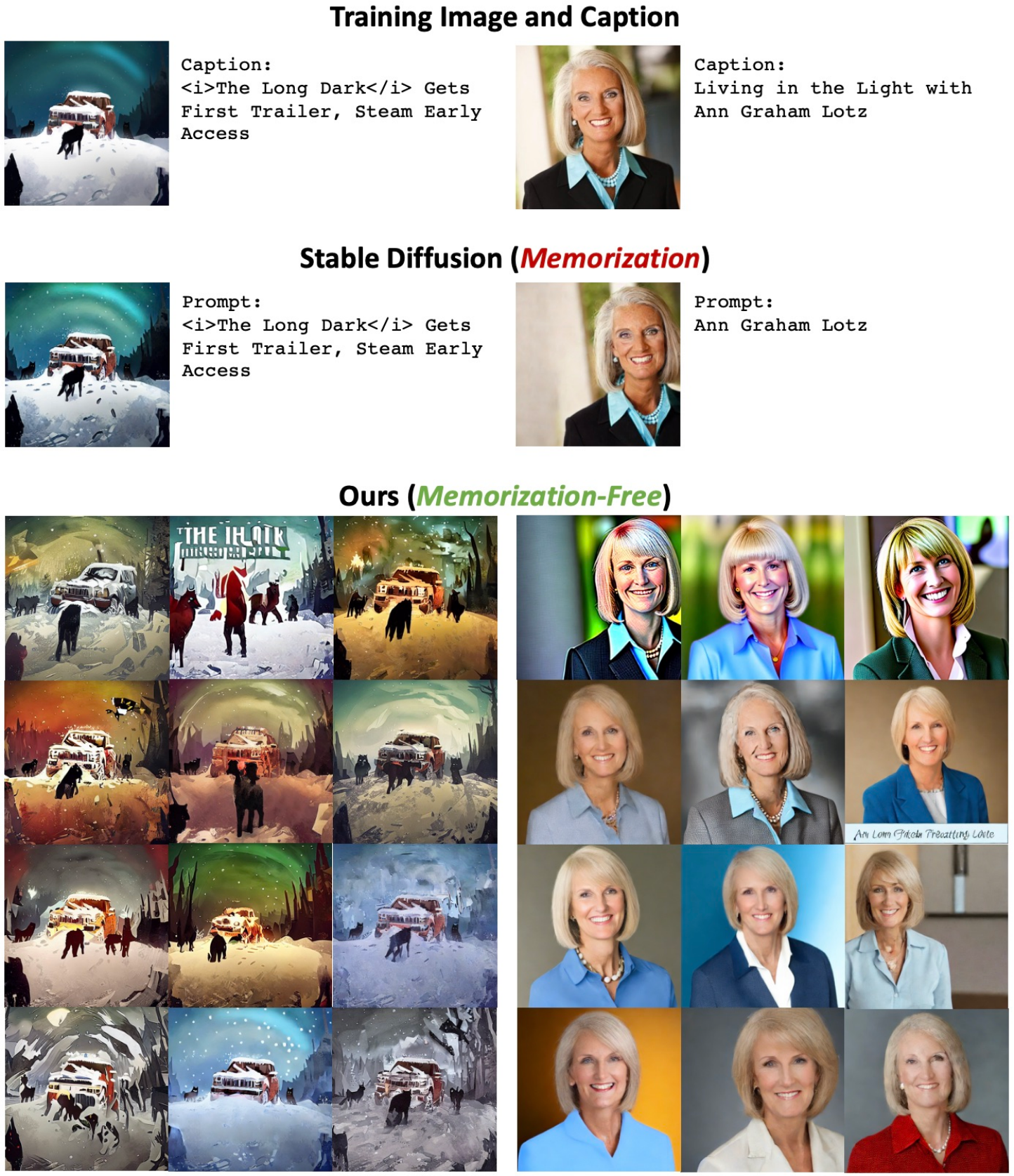}
  \captionsetup{font=large} 
  \caption{Additional qualitative comparisons showcase AMG's effectiveness in guiding pretrained diffusion models to produce memorization-free outputs.}
  \label{fig:qualitative_11_1}
\end{figure*}

\begin{figure*}[!ht]
  \centering
  \includegraphics[width=\textwidth]{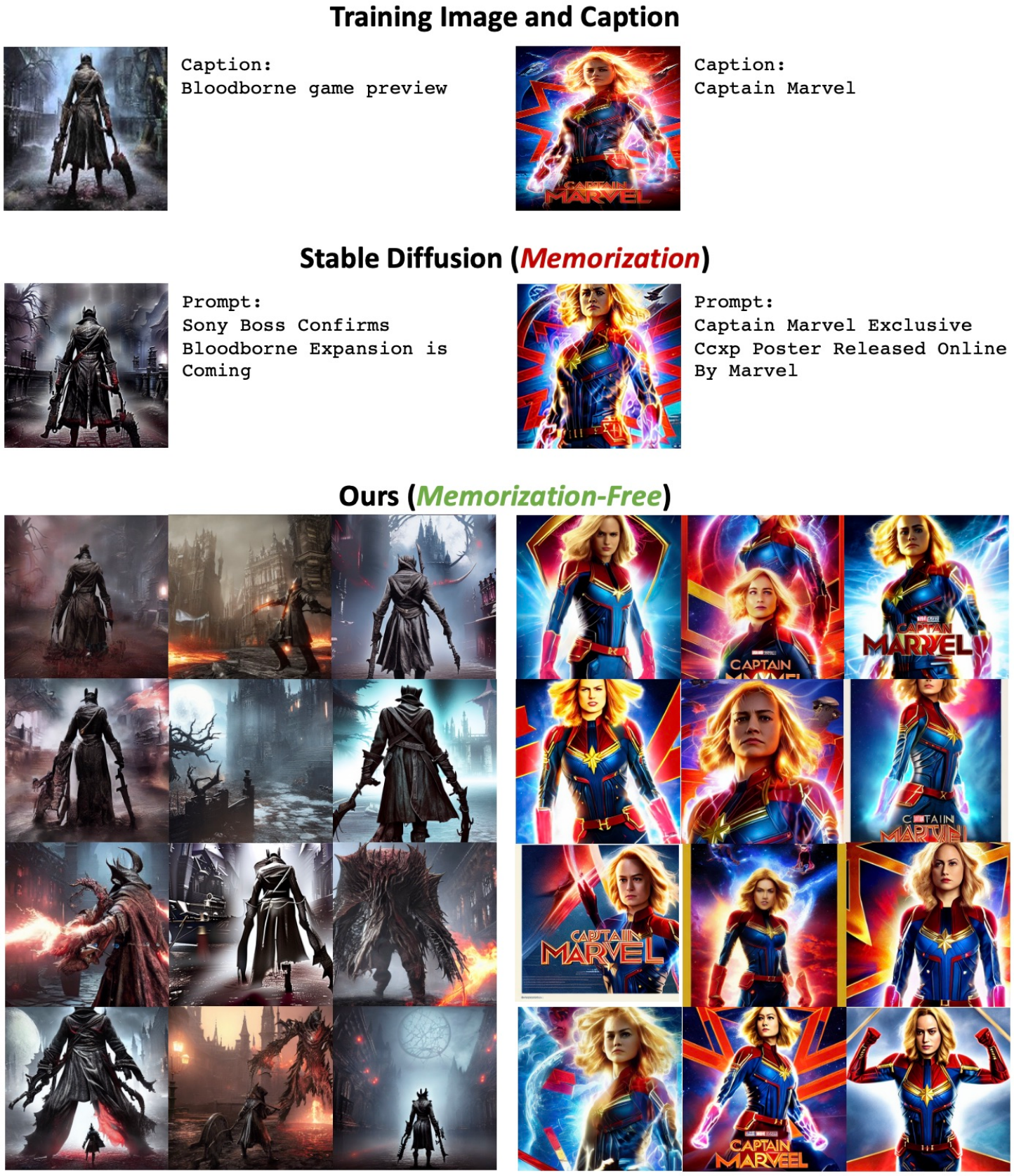}
  \captionsetup{font=large} 
  \caption{Additional qualitative comparisons showcase AMG's effectiveness in guiding pretrained diffusion models to produce memorization-free outputs.}
  \label{fig:qualitative10_9}
\end{figure*}

Finally, we present additional qualitative results in \cref{fig:qualitative8_6}, \cref{fig:qualitative_11_1}, and \cref{fig:qualitative10_9} to demonstrate AMG's effectiveness in guiding pretrained diffusion models to produce memorization-free outputs.

